\documentclass[10pt,journal]{IEEEtran}
\ifCLASSOPTIONcompsoc
  \usepackage[nocompress]{cite}
\else
  \usepackage{cite}
\fi

\usepackage[utf8]{inputenc} 
\usepackage[T1]{fontenc}    
\usepackage{hyperref}       
\usepackage{url}            
\usepackage{booktabs}       
\usepackage{amsfonts}       
\usepackage{nicefrac}       
\usepackage{microtype}      
\usepackage{lipsum}
\usepackage{graphicx}
\usepackage{multirow}
\usepackage{diagbox}

\usepackage{amssymb}
\usepackage{pifont}
\newcommand{\cmark}{\ding{51}}%
\newcommand{\xmark}{\ding{55}}%

\begin{document}

\title{Multi-View Video-Based 3D Hand Pose Estimation}

\author{Leyla~Khaleghi,
        Alireza~Sepas-Moghaddam,~\IEEEmembership{Member,~IEEE},
        Joshua~Marshall,~\IEEEmembership{Senior Member,~IEEE},
        Ali~Etemad,~\IEEEmembership{Senior Member,~IEEE}
\thanks{Leyla Khaleghi, Alireza Sepas-Moghaddam, Joshua~Marshall and Ali Etemad  are with the Department of Electrical \& Computer Engineering and the Ingenuity Labs Research Institute, Queen's University, Kingston, ON, Canada. Contact information: 19lk12@queensu.ca.}}
\IEEEtitleabstractindextext{%
\begin{abstract}
Hand pose estimation (HPE) can be used for a variety of human-computer interaction applications such as gesture-based control for physical or virtual/augmented reality devices. Recent works have shown that videos or multi-view images carry rich information regarding the hand, allowing for the development of more robust HPE systems. In this paper, we present the Multi-View Video-Based 3D Hand (MuViHand) dataset, consisting of multi-view videos of the hand along with ground-truth 3D pose labels. Our dataset includes more than 402,000 synthetic hand images available in 4,560 videos. The videos have been simultaneously captured from six different angles with complex backgrounds and random levels of dynamic lighting. The data has been captured from 10 distinct animated subjects using 12 cameras in a semi-circle topology where six tracking cameras only focus on the hand and the other six fixed cameras capture the entire body. Next, we implement MuViHandNet, a neural pipeline consisting of image encoders for obtaining visual embeddings of the hand, recurrent learners to learn both temporal and angular sequential information, and graph networks with U-Net architectures to estimate the final 3D pose information. We perform extensive experiments and show the challenging nature of this new dataset as well as the effectiveness of our proposed method. Ablation studies show the added value of each component in MuViHandNet, as well as the benefit of having temporal and sequential information in the dataset.
We make our dataset publicly available to contribute to the field at: {\href{https://github.com/LeylaKhaleghi/MuViHand}{https://github.com/LeylaKhaleghi/MuViHand}}.

\end{abstract}


}

\maketitle

\IEEEdisplaynontitleabstractindextext

%
\IEEEpeerreviewmaketitle


\section{Introduction}

Hand pose estimation (HPE) methods can play an important role in various human-computer interaction (HCI) applications, including virtual reality (VR) or augmented reality (AR) \cite{doosti2019hand, piumsomboon2013user, lee2009multithreaded, jang20153d,ahmad2019hand}, gesture and sign language recognition \cite{chang2016spatio, yin2016iterative,xu2021semi,do2020robust}, and smart vehicles \cite{de2019heterogeneous,rangesh2018handynet}. Despite the tremendous progress in HPE in recent years \cite{li2019survey,supanvcivc2018depth,chen2020survey} due to advancements in deep learning systems, the accuracy and robustness of HPE methods still suffer from:
(\textit{i}) appearance variations such as the articulated shape of the hand or skin color; (\textit{ii}) occlusion factors, such as wearing hand gloves or when a part of the subject's own body occludes the hand \cite{mueller2017real, ye2018occlusion}; (\textit{iii}) variations in the viewpoint of cameras \cite{simon2017hand}; and (\textit{iv}) variations in the environment, such as complex backgrounds \cite{mueller2018ganerated} and significantly high or low levels of lighting, which generally degrade the segmentation and estimation performance.

\begin{figure}[!t]
\includegraphics[width=0.5\textwidth]{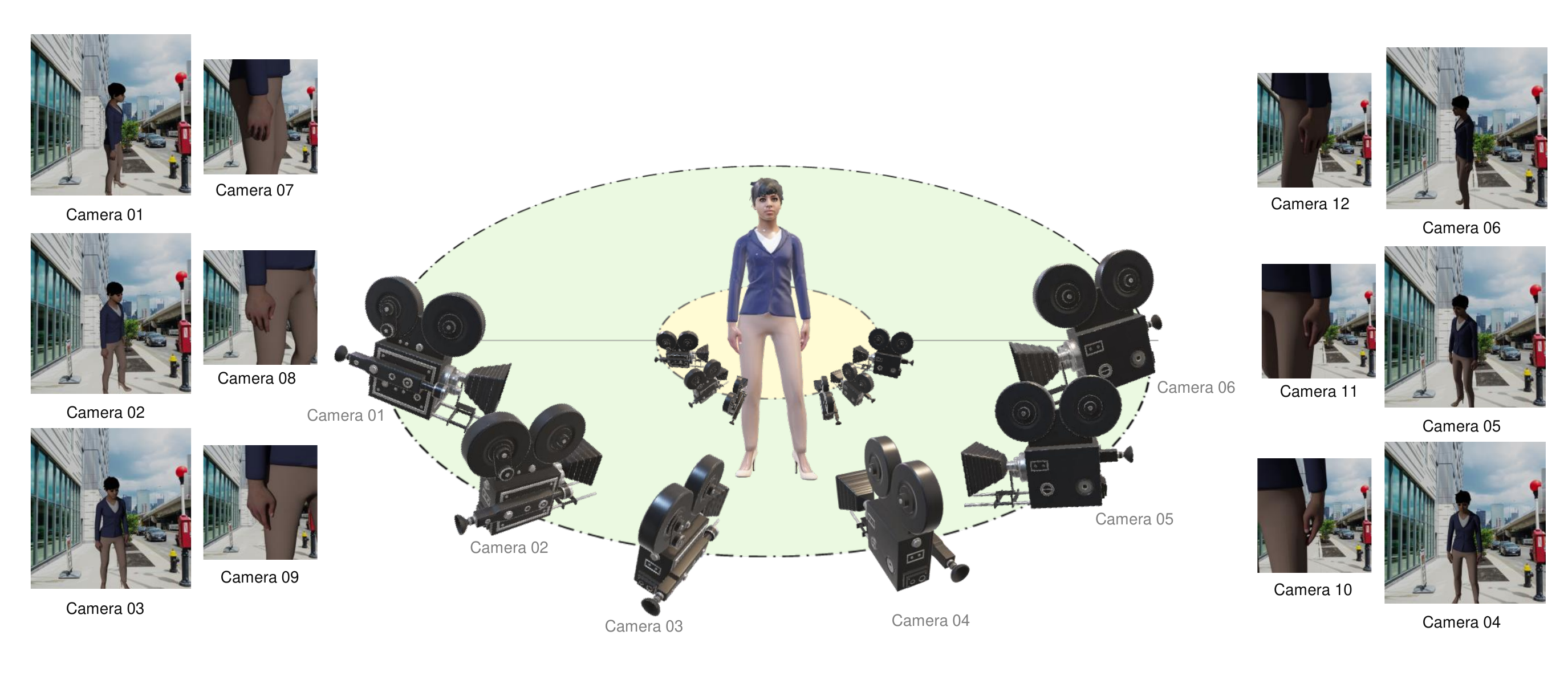}
\caption{The multi-view camera topology that has been used for recording the MuViHand dataset is presented in this figure. The videos have been recorded using 12 cameras situated in two concentric circles, where six tracking cameras (Cameras 1-6) focus only on the hand while the other six fixed cameras (Cameras 7-12) capture the whole body.}
\label{fig:dataset}
\end{figure}

\begin{table*}[!t]
\caption{Comparison between the MuViHand dataset and other public hand pose datasets.}
    \centering\begin{tabular}
    {| l | c c c c c c c c|}
    \hline
    Dataset & Year & RGB/Depth & Real/Synth. & \# Frames  & Static/Seq. & Multi-View & Full/Zoomed & Fixed/Tracking\\
    \hline
     \hline
     STB \cite{zhang20163d}&2016 & RGB+Depth &Real &18K&Seq. &No &Full &Fixed \\
     RHD \cite{zimmermann2017learning}&2017 & RGB+Depth &Synth  &43.7K &Static &No &Full  &Fixed \\
     GANerated \cite{mueller2018ganerated}&2018 & RGB &Synth  &330K &Static &No & Zoomed  &Fixed \\
    FreiHAND \cite{zimmermann2019freihand}&2019  &RGB    &Real   &134K&Static &No & Full &Fixed \\ 
    Youtube Hand \cite{kulon2020weakly}&2020 &RGB &Real&48.65K &Seq. &No &Full  & Fixed \\
    SeqHAND \cite{yang2020seqhand}&2020 &RGB &Synth   &410K &Seq. &No &Zoomed  & Fixed \\
    MVHM \cite{chen2021mvhm}&2020 &RGB+Depth &Synth  &320K &Static &Yes &Zoomed  &Fixed \\
    \hline
    MuViHand (ours) &2021  &RGB    &Synth &402K &Seq. &Yes &Both  & Both\\
    \hline
    \end{tabular}
    \label{tab:dataset}
\end{table*}

To date, most HPE solutions focus on estimating the pose from single RGB images \cite{zimmermann2017learning,iqbal2018hand,cai2018weakly,Yuan_2019_ICCV, kulon2020weakly, baek2019pushing,yang2019aligning,spurr2018cross,mueller2018ganerated,panteleris2018using,boukhayma20193d,hasson2019learning,yang2019aligning,theodoridis2020cross,gu20203d,dibra2018monocular}, and consequently less emphasis has been placed on exploiting multi-view and/or video data \cite{chen2021mvhm,yang2020seqhand}. Only a few methods have recently have made use of multi-view angular information \cite{chen2021mvhm}, whose fusion has shown to compensate for the shortage of data that a single view contains. Additionally, depth ambiguity can be significantly reduced by considering geometry relationships between viewpoints \cite{simon2017hand, he2020epipolar,chen2021mvhm}. Hence, multi-view pose estimation methods result in a better performance when compared with single view methods. Furthermore, hand poses generally change quickly and, because the pose at any given time can be influenced by the previous poses, exploiting temporal information over video sequences could boost the performance of HPE, as shown in \cite{yang2020seqhand, cai2019exploiting}.

Accordingly, the availability of large-scale video datasets captured from multiple view-points may play a key role in advancing the field of 3D HPE. Despite this, well-known hand-pose datasets such as RHD \cite{zimmermann2017learning}, GANerated \cite{mueller2018ganerated}, and FreiHAND \cite{zimmermann2019freihand} do not provide video data. Other datasets, like STB \cite{zhang20163d}, Youtube Hand \cite{kulon2020weakly}, and SeqHAND \cite{yang2020seqhand}, lack simultaneous recording of the hands from multiple views. Consequently, development of HPE solutions that can learn jointly from multi-view information over time (videos) has been largely overlooked.

To this end, we first introduce a new HPE dataset named \underline{Mu}lti-View \underline{Vi}deo-Based \underline{Hand} (MuViHand), to enable research on multi-view video-based HPE systems. Our dataset includes more than 402,000 synthetic hand images, available in 4,560 video sequences, which have been synthetically generated. The data has been captured by using 12 cameras in a semi-circle topology, where six tracking cameras focus only on the hand and the other six fixed cameras capture the whole body, as shown in Figure~\ref{fig:dataset}. To the best of our knowledge, MuViHand is the first and largest synthetic dataset that includes both multi-view and sequential hand data. Furthermore, the entire body is depicted in this dataset rather than images focusing only on the hand, causing additional challenges for HPE algorithms.

Motivated by the availability of the MuViHand dataset, we propose a graph-based HPE method, called MuViHandNet, to jointly learn from both temporal and angular information. Successive to extracting spatial embeddings from each frame using an encoder, 
our model uses a pair of temporal and angular learners to learn effective
spatio-temporal and spatio-angular representations. These representations are then concatenated and jointly learned for estimating 2D hand coordinates. Rather than estimating the 3D hand pose coordinates directly from the embeddings, MuviHandNet initially estimates the 2D hand coordinates prior to transforming them into 3D camera hand coordinates by implicitly estimating depth information. Given the graph-based structure of the hand skeleton, we use a graph convolutional network (GCN) to model the hand joint constraints and connections, and ultimately 3D HPE.



Our contributions in this paper are summarized as follows:

\begin{itemize}

\item We present the MuViHand dataset, which is the first and largest synthetic \textit{multi-view} \textit{video} hand pose dataset with two different types of cameras in which half capture the full body while the other half track the target hand. Our dataset includes more than 402,000 frames with complex backgrounds, occlusions, and dynamic sources of lighting. This dataset is publicly available to the research community.

\item We propose MuViHandNet, the first method for multi-view video-based 3D HPE that achieves a robust performance on our dataset by considering both the temporal and angular relationships between hand-image embeddings. Our model consists of an image encoder, temporal learners, angular learners, and a graph U-Net.

\item We demonstrate the benefits of this new multi-view video-based dataset and 3D HPE model, and present a comprehensive benchmarking study against other state-of-the-art HPE methods as well as ablated baselines. The experiments show that MuViHandNet achieves superior results, with considerable performance gains of up to 55\% when compared to other state-of-the-art methods.

\end{itemize}

The remainder of this paper is organized as follows. Section 2 provides a review of recent advances in HPE datasets and methods. The newly captured hand dataset is described in Section 3 and the proposed method is presented in Section 4. Section 5 presents an extensive performance evaluation for the proposed and state-of-the-art methods using varied and challenging HPE tasks. Also, the limitation of our method with some remarks for future research directions are discussed. Finally, Section 6 concludes our paper.
 


\section{Related Work}
\subsection{Existing Hand Pose Datasets}
Several publicly available hand pose datasets have been previously developed for different applications and scenarios. Table~\ref{tab:dataset} provides an overview of the main characteristics of the well-known hand pose datasets, including the types of cameras used, the types of data (real vs. synthetic and static vs. sequential), the number of frames, the type of view-related acquisition (multi-view or not), whether the images only capture the hand or whether the entire body has been captured (which is more challenging), and whether the employed cameras are fixed or track the hand. In order to show the chronological evolution of these hand pose datasets in the table, they have been sorted by the order of release date.  For comparison, we also included the characteristics of our dataset, MuViHand, which is proposed in this paper in Table~\ref{tab:dataset}. We note that MuViHand is among those datasets with the highest number of frames and is both multi-view and sequential, which no other dataset provides. Moreover, MuViHand contains the entire body, which is more representative of real-life scenarios while also making it more challenging for pose estimators because other body parts with similar skin characteristics might confuse candidate algorithms. Lastly, our dataset is the only dataset that contains both fixed and tracking cameras. In what follows we provide a brief description of the datasets mentioned in Table~\ref{tab:dataset}.

\begin{figure*}[!t]
\centerline{\includegraphics[width=1\textwidth]{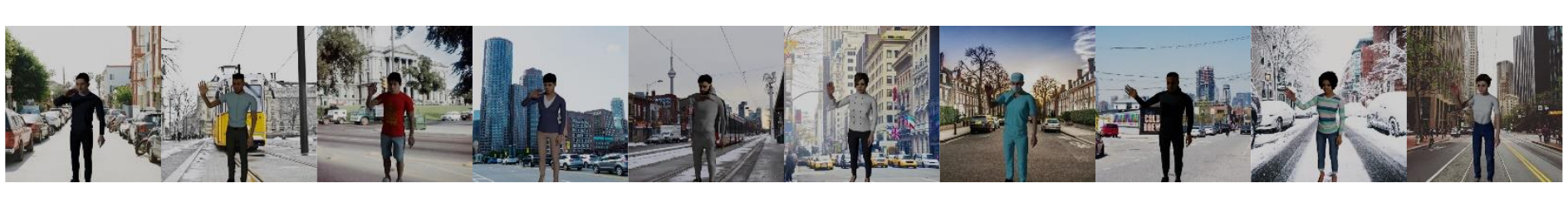}}
\caption{Illustration of each of the animated subjects in MuViHand, showing different genders, skin tones, and appearances.}
\label{fig:subject}
\end{figure*}

\textbf{Stereo Hand Pose Tracking Benchmark (STB)} \cite{zhang20163d} is one of the most popular real-world single-view datasets for 3D HPE. In this dataset, only one subject perform random and number counting poses with six different backgrounds. These 12 sequences, each with 1500 frames, were annotated manually. 

\textbf{Rendered Hand Pose Dataset (RHD)}\cite{zimmermann2017learning} is a synthetic hand pose dataset, including 20 different characters who perform 39 actions. The dataset includes 43,700 images that were captured with multiple random backgrounds from different angles (not recorded simultaneously, hence not multi-view). Given that the dataset uses synthetic images, the keypoints were annotated automatically.

\textbf{GANerated}\cite{mueller2018ganerated} was first generated synthetically, and subsequently used by a CycleGan network and translated to real images. GANerated includes more than 260,000 frames of hand poses with different skin tones.

\noindent \textbf{FreiHAND}\cite{zimmermann2019freihand} is a large-scale real-world hand pose dataset collected from 32 subjects. The dataset includes 134,000 hand images with various poses. The annotations were partially carried out manually and partially by an iterative semi-automated approach.

\textbf{YouTube Hand} \cite {kulon2020weakly} was generated from 109 YouTube videos. The dataset includes 48,650 hand images. The process of annotation was done by fitting a parametric hand model, named MANO\cite{romero2017embodied}, to a publicly available 2D prediction network called OpenPose \cite{simon2017hand}.

\textbf{SeqHAND} \cite {yang2020seqhand} is a large synthetic sequential hand pose dataset with 410,000 images. The MANO \cite{romero2017embodied} hand model was used with the ground-truth annotations of a sequential depth-based hand pose dataset called BigHand 2.2M, to render the hand pose sequences for the SeqHAND dataset.

\textbf{Multi-View Hand Mesh (MVHM)} \cite{chen2021mvhm} is a synthetic multi-view dataset with 320,000 images. Similar to the previous dataset, MVHM used ground truth annotations from a different dataset, in this case the NYU \cite{tompson2014real}, along with a hand model called TurboSquid\footnote{Available online at \href{https://www.turbosquid.com/}{https://www.turbosquid.com}.}, to render the hand images from eight views.

\subsection{Existing HPE Methods}
While in this paper we focus on {\em multi-view} and {\em video-based} HPE, in this section we review related literature that has performed 3D HPE from single RGB images given the lack of multi-view and video-based datasets \cite{zimmermann2017learning,iqbal2018hand,cai2018weakly,Yuan_2019_ICCV, kulon2020weakly, baek2019pushing,yang2019aligning,spurr2018cross,mueller2018ganerated,panteleris2018using,boukhayma20193d,hasson2019learning,yang2019aligning,theodoridis2020cross,gu20203d}. We then review the few works that have taken multi-view or sequential approaches \cite{yang2020seqhand,simon2017hand, chen2021mvhm}. 

\subsubsection{Single-view HPE}
In \cite {zimmermann2017learning}, the problem of 3D HPE is broken up into two steps. First, a CNN extracts image features and directly estimates the 2D heat maps, from which the normalized 3D hand coordinate is subsequently measured. Similar to \cite {zimmermann2017learning}, in \cite{iqbal2018hand}, 3D HPE is performed in two steps. However, the 3D hand pose is estimated from a 2.5D heat map instead of a 2D heat map by estimating the depth data as well.

Some HPE methods \cite{cai2018weakly,Yuan_2019_ICCV} boost the performance of RGB-based HPE with the help of privileged learning of depth information. In \cite{cai2018weakly}, a depth regularizer network is applied after the 3D HPE network during training to learn to generate the corresponding depth map from a 3D hand pose. However, during testing the RGB images go through only the 3D HPE network. Similarly, in \cite{Ge_2019_CVPR} the network learns to generate the corresponding depth map from the 3D hand shape instead of the pose. In \cite{Yuan_2019_ICCV}, an RGB-based HPE and a depth-based HPE network are independently trained. The depth-based network is then frozen and the RGB-based network's training is resumed with paired RGB and depth images by sharing the information between the middle CNN layers of these two networks.

Several HPE methods \cite{panteleris2018using,boukhayma20193d,hasson2019learning} rely on a predefined 3D hand model for estimating 3D hand poses. In \cite{panteleris2018using}, with the help of solving an optimization problem, a hand model with 27 parameters are fit to 2D joints locations estimated by OpenPose \cite{simon2017hand}. In \cite{hasson2019learning} a hand image is passed through a ResNet-18 to generate the MANO hand model parameters for estimating the 3D hand pose. Similarly, in \cite{boukhayma20193d}, a hand image and its 2D joint heat maps (obtained from OpenPose) are passed through a ResNet-50 for generating the input parameters to the MANO hand model, from which the 3D hand pose is measured.

\subsubsection{Multi-View HPE}
In the only multi-view HPE method found in the literature, the proposed solution \cite{chen2021mvhm} receives a number of hand images captured from different views as inputs, and passes them to individual single-view HPE networks to predict the 3D camera coordinates for each view independently. Then these 3D camera coordinates are concatenated and pass through a graph-based neural network to predict the 3D \textit{world} coordinates for each pose.

\subsubsection{Temporal HPE}

In \cite{cai2019exploiting} 3D hand pose is estimated from a temporal sequence of 2D hand joints obtained from a hand pose estimator such as OpenPose. The method creates a GCN that considers the temporal relationships by incorporating additional edges between the same joints in consecutive frames. In \cite{yang2020seqhand}, a recurrent layer is included after the encoder of a single-view hand pose estimator (proposed in \cite{boukhayma20193d}) to exploit the temporal relationships and generate the MANO hand model parameters for each frame.

Based on our above-described literature review, we observe that there are no HPE methods that simultaneously consider both spatio-temporal and spatio-angular relationships. Thus, this paper introduces the first method for multi-view video-based HPE, called MuViHandNet.

\begin{figure}[!t]
\centerline{\includegraphics[width=0.5\textwidth,height=0.5\textwidth]{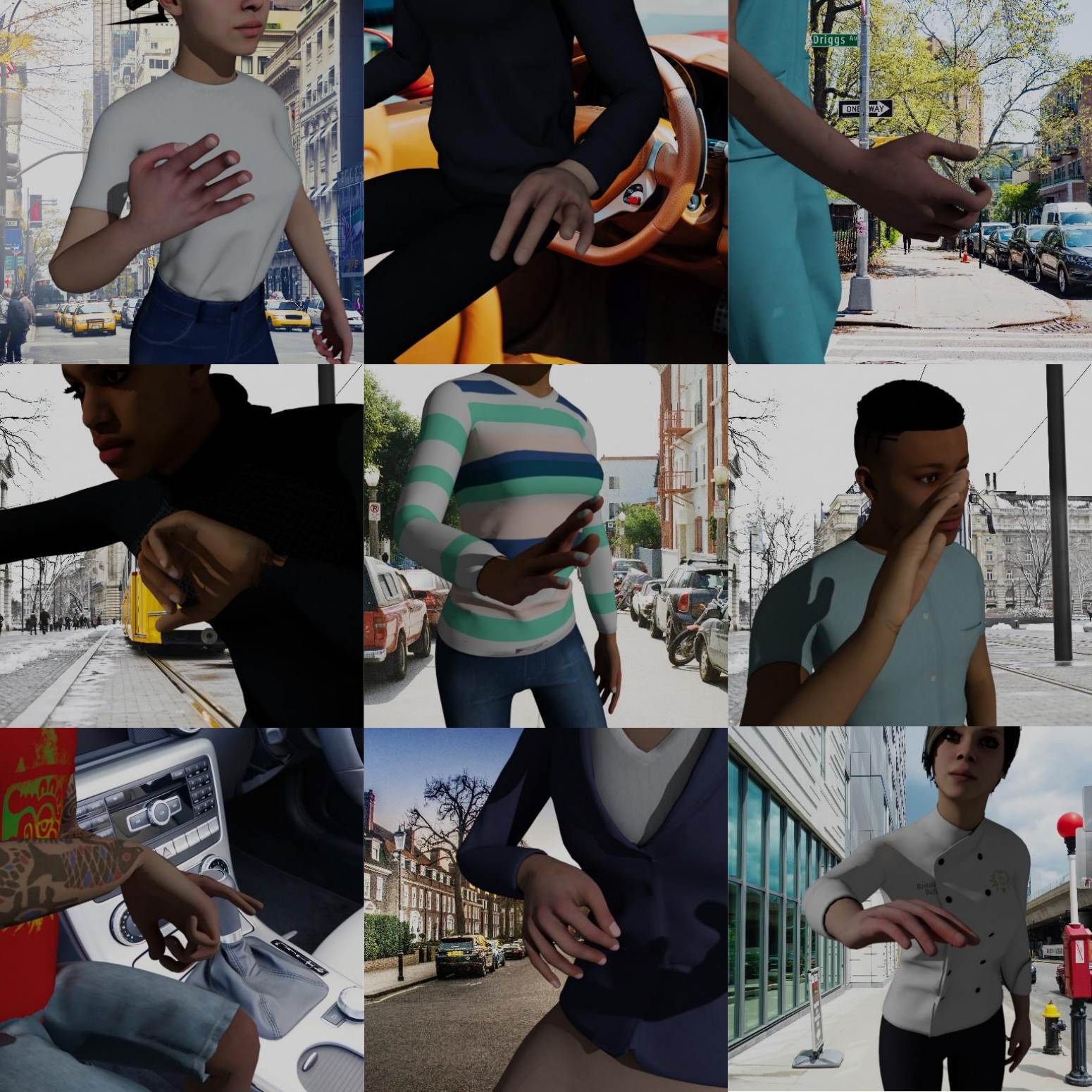}}
\caption {Samples from MuViHand with various poses and backgrounds.}
\label{fig:dataset_sample}
\end{figure}

\begin{figure}[!t]
\centerline{\includegraphics[width=0.5\textwidth]{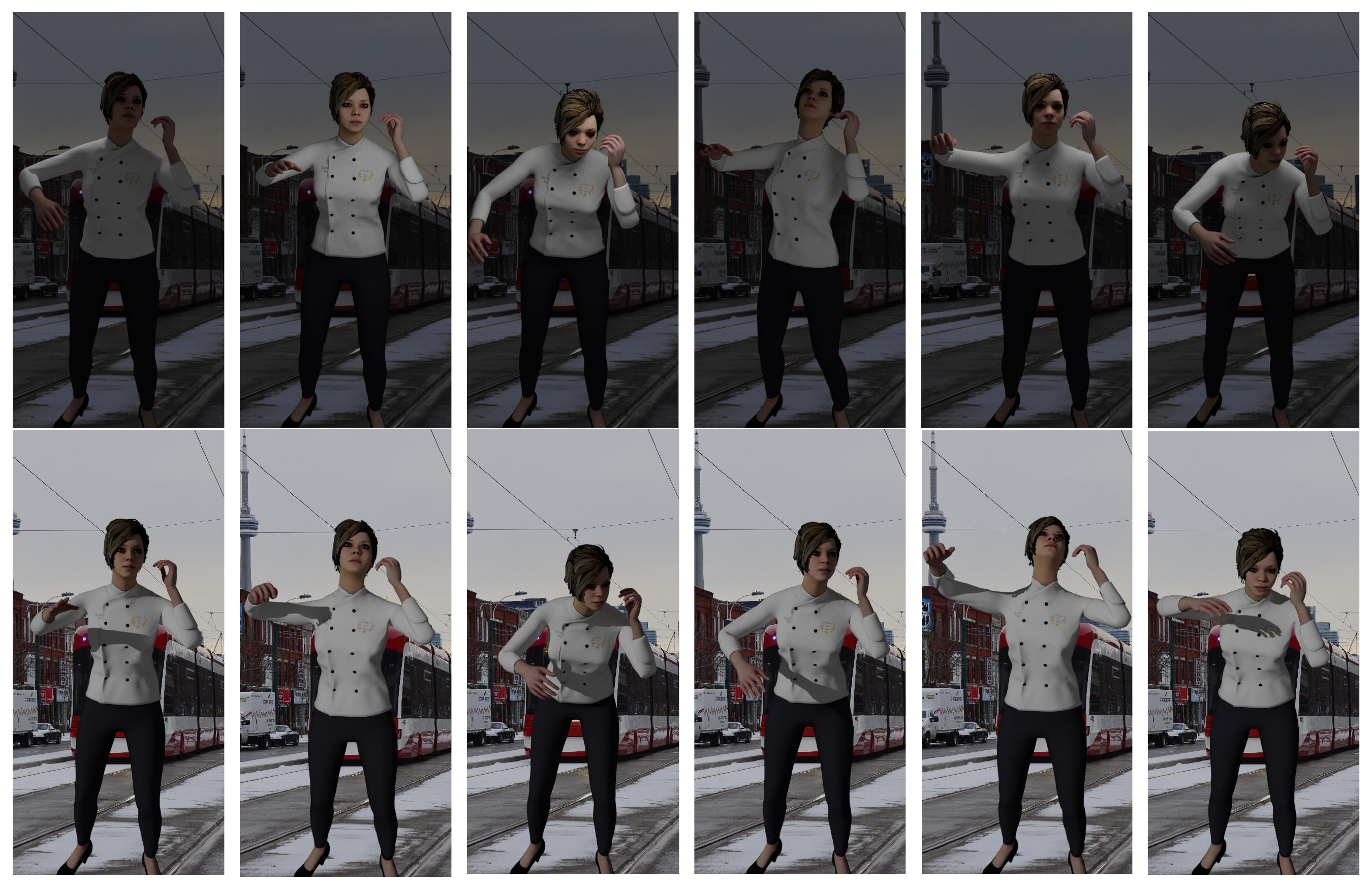}}
\caption {Samples with different light sources. In the first row the light source is a point light and in the second row the light source is the sun. }
\label{fig:light_variation}
\end{figure}
\begin{figure}[!t]
\centerline{\includegraphics[width=0.9\columnwidth]{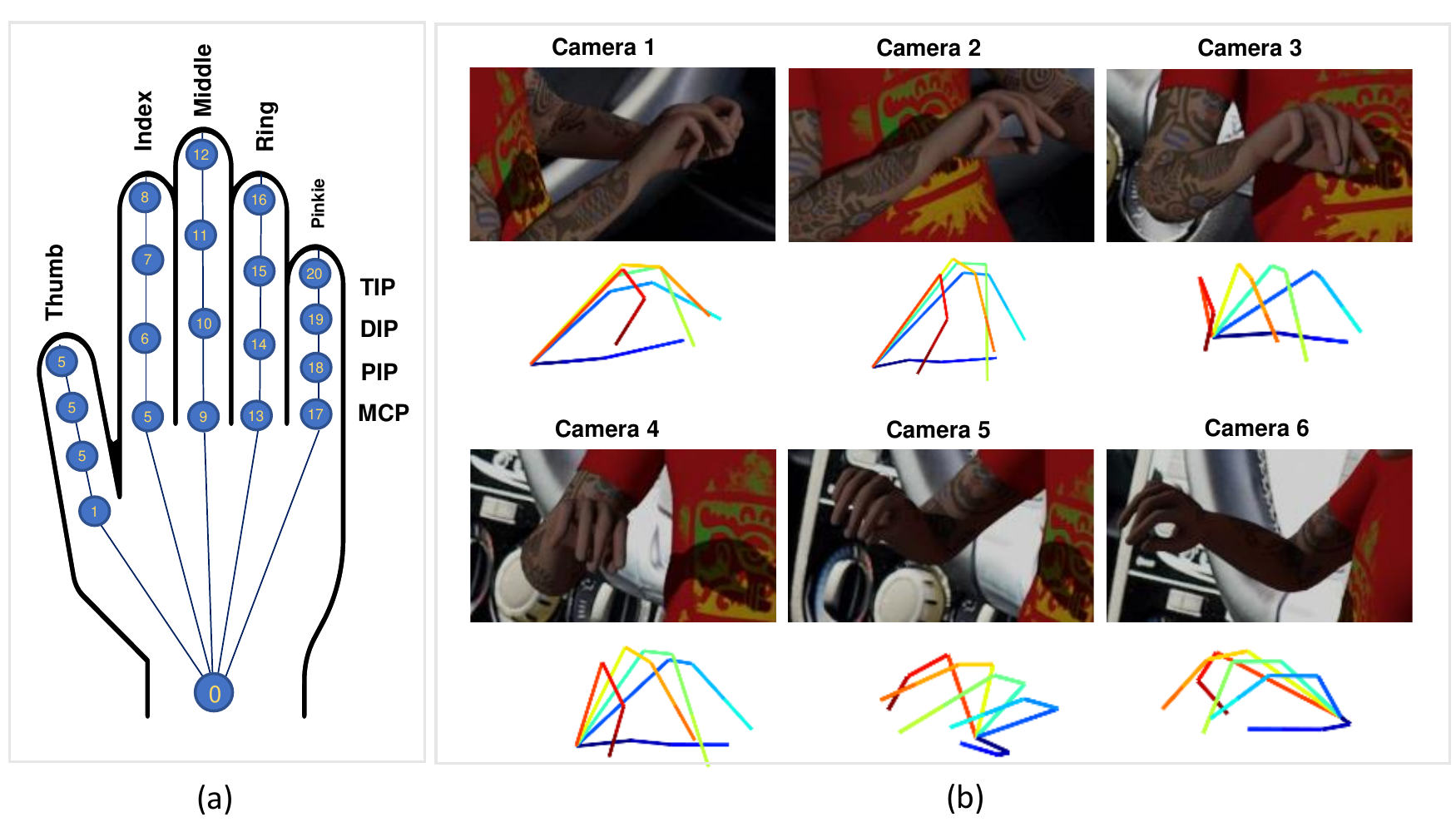}}
\caption{(a) The 21 hand joints are visualized; (b) A sample hand pose captured from cameras 1 through 6, and their corresponding 3D ground truths are depicted.}
\label{fig:6view}
\end{figure}

\section{Dataset}
Our proposed dataset, MuViHand, is a synthetic multi-view video-based hand pose dataset, create using the freely available MIXAMO\footnote{Available online at \href{https://www.mixamo.com}{https://www.mixamo.com}.}, a web-based service for 3D character animation synthesis. MIXAMO has also been used for creating the Rendered Hand Pose Dataset \cite{zimmermann2017learning}. In our work, we select 10 characters with various appearances (shown in Figure~\ref{fig:subject}) who perform 19 various full-body actions in sitting, standing, and walking posture, that have been captured from motion actors. We also randomly choose several street and in-vehicle images as backgrounds from the Pxfuel\footnote{Available online at \href{https://www.pxfuel.com}{https://www.pxfuel.com}.} website. Finally, Blender\footnote{Availabel online at \href{https://www.blender.org/}{https://www.blender.org}.}, an open-source 3D graphic computer software, is used to render the videos from six different views. In total, 4,560 videos with 402,000 frames are rendered, a few samples of which are shown in Figure~\ref{fig:dataset_sample}. This dataset is publicly available at \href{https://github.com/LeylaKhaleghi/MuViHand}{https://github.com/LeylaKhaleghi/MuViHand}.

\subsection{Lighting Conditions}
We use two dynamic lighting sources in creating the videos, namely (\textit{i}) sun and (\textit{ii}) point light. In Figure~\ref{fig:light_variation} several frames with the two sources of light are shown. In the videos with sun as the light source, the location of the sun is randomly picked between two concentric semi-spheres around the subject. Moreover, the sun source experiences very small motions towards the left or right during each video to create a slightly moving shadow.
The point light sources are similarly located between two concentric spheres with much smaller radii compared to those of the sun light. During each video, the point light source moves closer to the user, creating a moving shadow similar to that experienced as a result of a moving vehicle at night.

\subsection{Camera Topology}
We use 12 cameras when generating the dataset, six of which are \textbf{fixed} in a semi-circle topology around the user, while the other six  \textbf{track} the user, three of which focus on the right hand 
while the other three focus on the left hand.
This topology is shown in Figure~\ref{fig:dataset}.
Each set of cameras (fixed and tracking) have been positioned in six evenly spaced angles from $15^\circ$ to $165^\circ$ ($15^\circ$, $45^\circ$, $75^\circ$, \dots, $165^\circ$) on a semicircle topology around the subject. Figure~\ref{fig:6view} presents a sample hand image and corresponding ground-truth poses as observed by the six fixed cameras.

\subsection{Annotation}
We provide 2D and 3D locations for 21 hand joints, including one joint for the wrist and four joints per each finger (Fingertip, DIP, PIP, MCP), similar to \cite{zimmermann2017learning}. See Figure \ref{fig:6view}(a) for details about the joints used in this dataset. Moreover, Figure \ref{fig:6view}(b) shows sample ground-truth poses provided for different views.
For each 3D hand pose, we provide the 3D world coordinates as well as the 3D camera coordinates. Finally, the intrinsic matrices for the cameras are also provided, where a single intrinsic matrix describes each static camera, while for the tracking cameras, a different matrix is provided for each frame.

\subsection{Pose and Activity Distribution}
\begin{figure}[!t]
\centerline{\includegraphics[width=0.9\columnwidth]{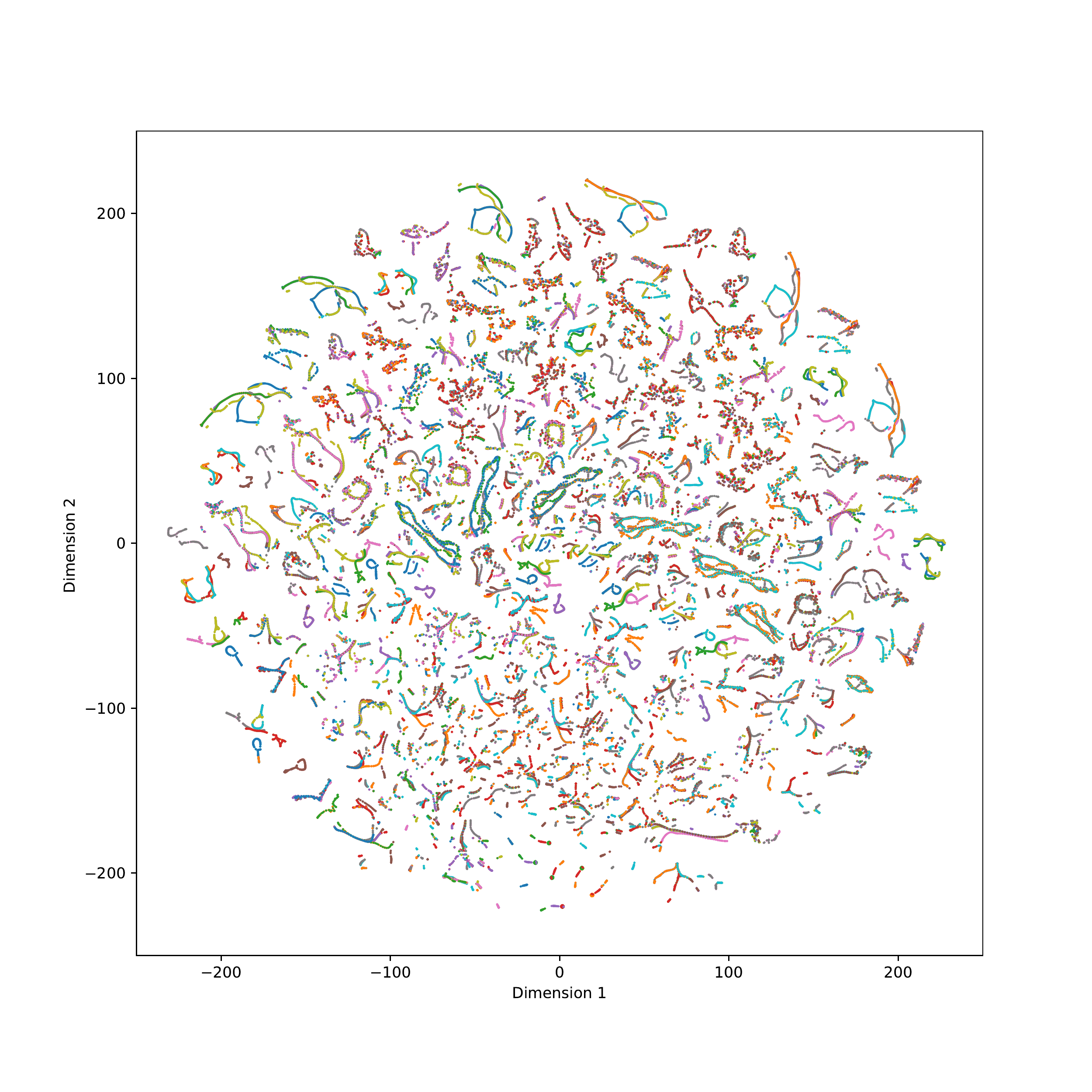}}
\caption{t-SNE visualization for the pose-activity space in the dataset. 
Each sample is a 3D root-relative pose ($1 \times 63$ vector) of the right hand captured from Cameras $1,2, \cdots,6$. Each color represents a distinct activity.}
\label{fig:TSNE}
\end{figure}

In order to illustrate how the pose information is distributed in our dataset, we visualize the 3D root-relative pose coordinates ($1 \times 63$ vector) using t-Distributed Stochastic Neighbor Embedding (t-SNE) in Figure~\ref{fig:TSNE}, similar to \cite{garcia2018first,yang2020seqhand,yuan2017bighand2}. We observe that the pose space captured by the dataset is quite varied and non-skewed. Moreover, we use different colours to represent each of the 19 activities used in the dataset. From the figure, we observe that the distribution of activities is also spread out in the pose space.

\section{Method}\label{sec:Method}


Our work builds upon the idea that learning temporal \cite{simon2017hand, he2020epipolar,chen2021mvhm} or angular \cite{yang2020seqhand, cai2019exploiting} information is beneficial for HPE. We propose MuViHandNet, $\xi$, a deep neural network for predicting the 3D hand camera coordinates $\textit{P}$ from the corresponding \textit{multi-view videos} $\Phi$, such that 
\begin{equation}
    \textit{P}= \xi[\Phi], 
\end{equation}
where $\Phi = \{\phi_{v}^ {t}\}$ is a frame captured at time $t$ and from view $v$. Each hand frame is described by an RGB image $\phi_{v}^ {t} \in \mathbb{R}^{{3} \times{H} \times{W}}$, where $H$ is the height of the image and $W$ is the image's width. Moreover, the 3D hand camera coordinates at time $t$ and view $v$, $P = \{p_{v}^ {t}\} \in \mathbb{R}^{{j} \times {3}}$, are described by a hand skeleton with $J$ joints.

\begin{figure*}[!t]
\centering
\includegraphics[width=0.95\textwidth]{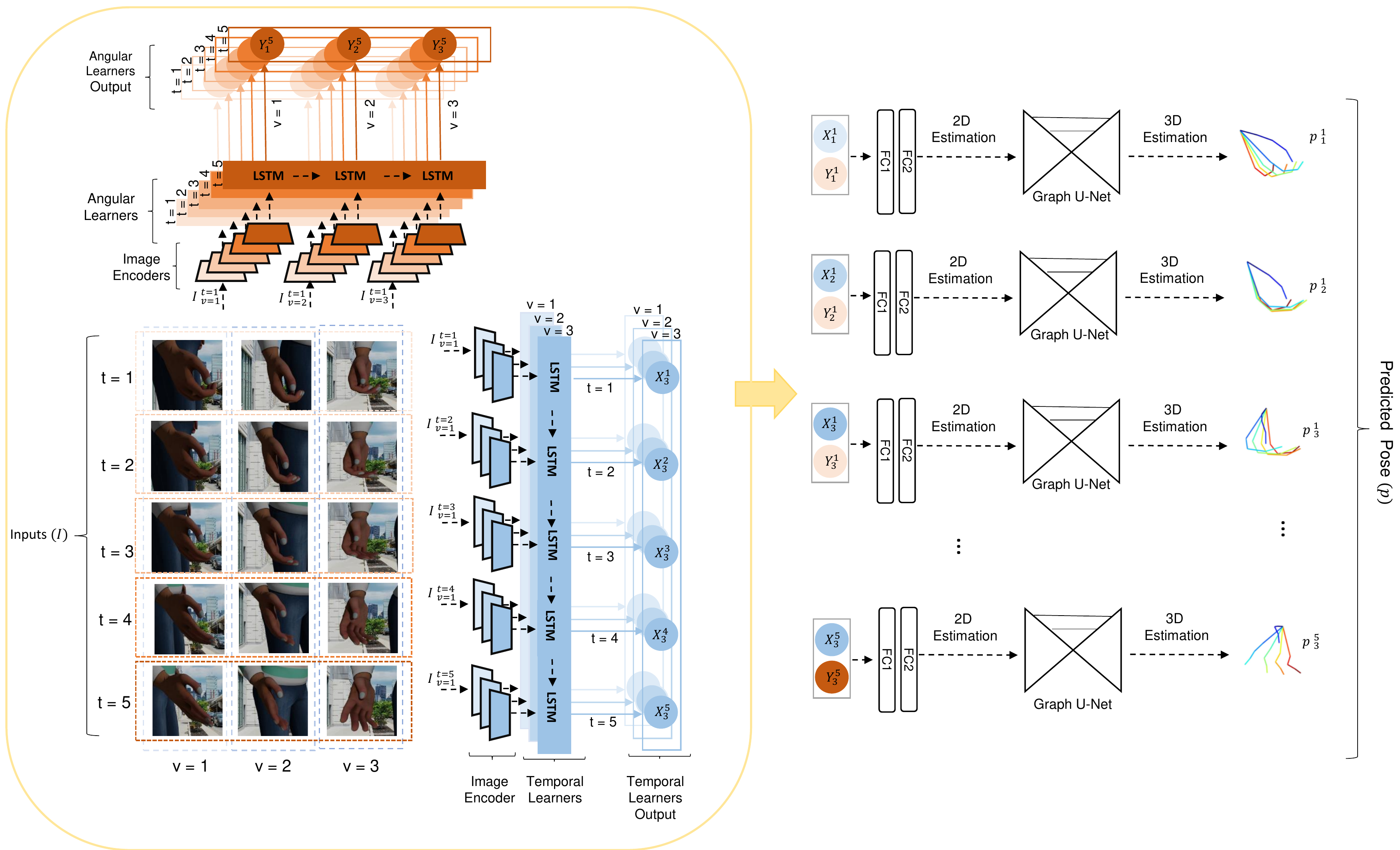}
\caption{The pipeline for MuViHandNet is depicted. The model consists of image encoders, temporal learners (LSTM$_t$), angular learners (LSTM$_v$), and graph U-Nets. Each LSTM$_t$, takes 3 hand videos from $v=1$ to $v=3$ for all 5 frames, and generates a feature vector, ${X_v^t}$, per each frame. Furthermore, LSTM$_v$ takes 5 hand sequences in different time steps $t=1$ to $t=5$ for all 3 frames, and generates a feature vector, ${Y_v^t}$, for each frame. The concatenation of ${X_v^t}$ and ${Y_v^t}$ for each frame is fed through fully connected layers to estimate the 2D coordinates, which are then passed through the graph U-Net to predict the 3D hand pose, ${p}_{v}^ {t}$, for each frame.}
\label{fig:main_model}
\end{figure*}

\begin{figure}[!t]
\centerline{\includegraphics[width=0.45\textwidth]{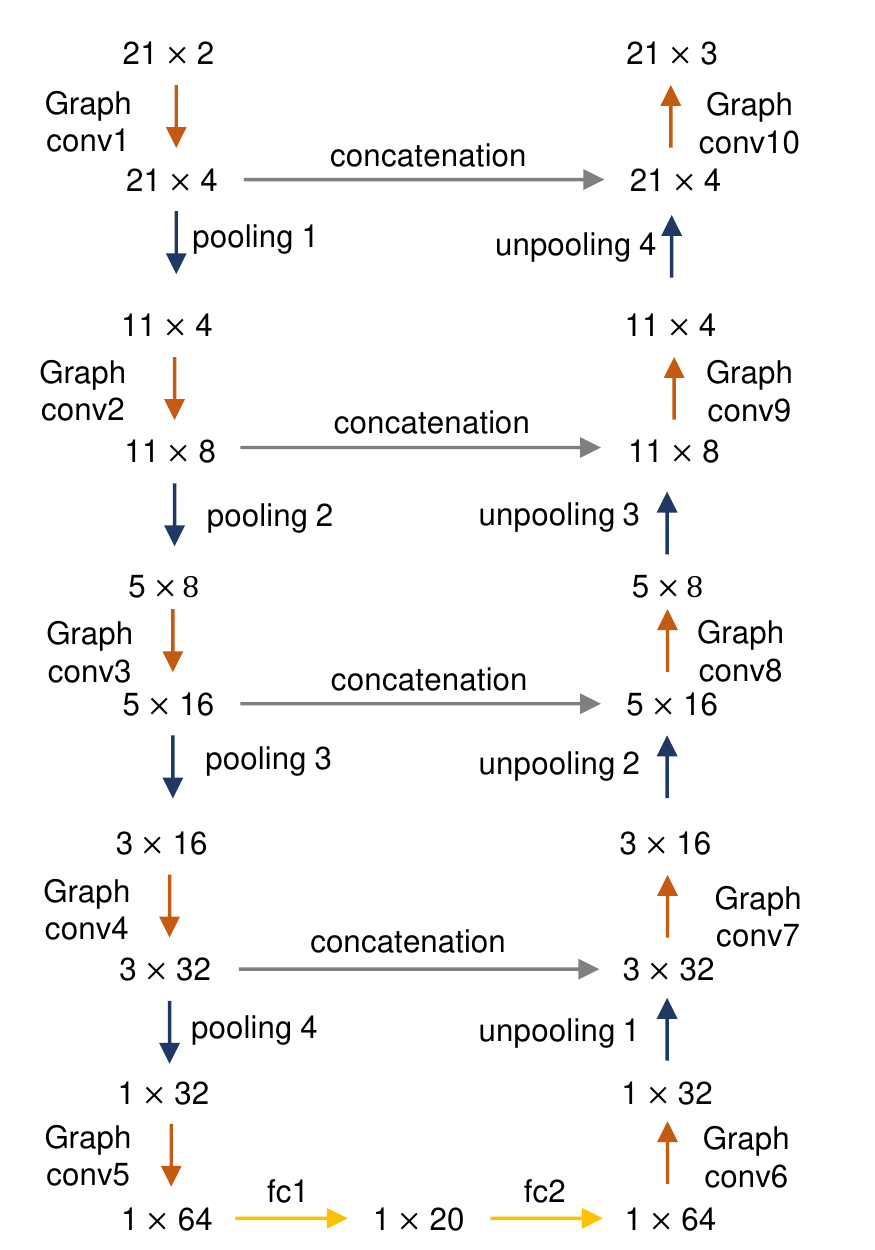}}
\caption {The graph U-Net architecture is presented. 
}
\label{fig:graphUnet}
\end{figure}

\subsection{Model Overview}

Figure \ref{fig:main_model} depicts the overview of our proposed network architecture. Similar to many state-of-the-art HPE methods \cite{ng2021body2hands,hasson2019learning,boukhayma20193d,doosti2019hand,yang2020seqhand,zimmermann2019freihand}, we use a ResNet as the encoder to extract features from each frame following by recurrent neural networks (RNNs) to 
learn the temporal relationships among the embeddings of the video frames. 
Furthermore, as shown in prior literature (in other domains) that RNNs can also effectively learn \textit{angular} relationships \cite{sepas2020facial,sepas2019deep,sepas2020long,sepas2021multi}, we incorporate an additional RNN in our model. The outputs of the two sets of RNNs (temporal and angular) are combined and used to estimate 2D joint positions, which are then fed to a Graph U-Net to provide the final 3D hand pose. Two-stage estimation of 3D hand pose (2D followed by 3D) has been widely used in prior works and shown advantages over direct 3D HPE \cite{iqbal2018hand,zimmermann2017learning,mueller2018ganerated,doosti2020hope}.

\subsection{Image Encoder}
We use a ResNet10 \cite{he2016deep} pre-trained on ImageNet \cite{russakovsky2015imagenet}, as the image encoder. This encoder takes each hand frame $\phi_{v}^ {t}\ $ as its input and generates an embedding $F_{v}^ {t}$ at time $t$ and from view $v$, where $t  = 1,2, \cdots ,T$, $T$ is the video length, $v = 1,2, \cdots ,V$, and $V$ is the number of the views.

\subsection{Temporal and Angular Learning}
Long short-term memory (LSTM) networks \cite{hochreiter1997long} are a popular type of RNNs used in a variety of different applications \cite{lu2017online,huang2020lstm,zhang2018attention}. Due to their effective performance in many domains, we use LSTMs for both spatio-temporal and spatio-angular learning. Each LSTM unit consists of three different gates, namely input ${r}$, forget ${f}$, an output ${o}$ gates as well as cell ${c}$ and hidden ${h}$ memories. Here we introduce the LSTM equations for the $k^{th}$ instance of a sequence, where `instance' is defined as a frame in a video or a particular viewpoint in a multi-view sequence.

Initially, the input gate for the $k^{th}$ instance of a sequence is computed according to 
\begin{equation}\label{eq:eqinput}
    \textit{r}_k= \sigma( \textit{w}_{rx}\textit{x}_k + \textit{w}_{rh}\textit{h}_{k-1}+ \textit{b}_r),
\end{equation}
where ${x}_k$ is the input vector to the LSTM unit, ${h}_{k-1}$ is the previous hidden state, ${w}_{rx}$ and ${w}_{rh}$ are the input gate weights, and ${b}_r$ is the input gate bias. $\sigma$ denotes the sigmoid activation.  In order to  control how the cell forgets information from its state, the forget gate ${f}_k$ is computed according to
\begin{equation}\label{eq:eqforget}
    \textit{f}_k= \sigma( \textit{w}_{fx}\textit{x}_k + \textit{w}_{fh}\textit{h}_{k-1}+ \textit{b}_f),
\end{equation}
where ${w}_{fx}$ and ${w}_{fh}$ are the forget gate weights, and ${b}_f$ is the forget gate bias. The output is then computed by using 
\begin{equation}\label{eq:eqout}
    \textit{o}_k= \sigma( \textit{w}_{ox}\textit{x}_k + \textit{w}_{oh}\textit{h}_{k-1}+ \textit{b}_o),
\end{equation}
where ${w}_{ox}$ and ${w}_{oh}$ are the output gate weights, and ${b}_o$ is the output gate bias.

The cell state that controls remembering values over the sequence (time or view) is updated according to 
\begin{equation}\label{eq:equC1}
    \tilde{\textit{c}}_k= \tanh( \textit{w}_{cx}\textit{x}_k + \textit{w}_{ch}\textit{h}_{k-1}+ \textit{b}_c),
\end{equation}
and
\begin{equation}\label{eq:equC2}
    \textit{c}_k = \textit{i}_k \odot \tilde{\textit{c}}_k + \textit{f}_k \odot \textit{c}_{k-1},
\end{equation}
where ${w}_{cx}$ and ${w}_{ch}$ are the cell weights, and ${b}_c$ is the cell bias. Finally, the hidden state ${h}_k $, is computed based on learning jointly the cell state ${c}_{k}$, and the output gates ${o}_k$ according to
\begin{equation}
    \textit{h}_k = \textit{o}_k \odot \tanh(\textit{c}_{k}).
\end{equation}

As mentioned earlier, we use two sets of LSTM networks to learn the temporal and angular relationships separately, which we name the temporal learner and angular learner, respectively. We denote these networks
LSTM$_t$ and LSTM$_v$. 
Accordingly, 
\begin{equation}
[X^1_v, X^2_v,\cdots, X^T_v] = LSTM_t([F^1_v, F^2_v,\cdots, F^T_v]), 
\end{equation}
and
\begin{equation}
[Y^t_1, Y^t_2,\cdots, Y^t_V] = LSTM_v([F^t_1, F^t_2, \cdots, F^t_V]),
\end{equation}
where $X^t_v$ and $Y^t_v$ are the outputs of the temporal and angular learner cells at time $t$ and view $v$.

Next, for the temporal and angular information to jointly contribute to the final HPE, the two sets of LSTM outputs $X^t_v$ and $Y^t_v$ are concatenated by using
\begin{equation} 
Z^t_v = X^t_v \oplus Y^t_v,
\end{equation}
where $\oplus$ denotes the concatenation operation and $Z_v^t$ is the joint feature set. In total, a set of features with $V \times T$ vectors are generated at this stage. Each $Z_v^t$ of the $V \times T$ vectors are then fed to 2 fully connected layers to produce the 2D coordinates ($w^t_v$) for each frame. 

\subsection{Graph U-Net}\label{sec:graohUnet}
Because the goal of this work is to perform 3D HPE, the 2D coordinates estimated by the FC layers following the temporal and angular learners need to be converted to 3D. In this context, it has been extensively shown in the literature \cite{doosti2020hope,chen2021mvhm,Ge_2019_CVPR,cai2019exploiting} that the graph-based structure of the hand skeleton lends itself well to GCN-style networks \cite{kipf2016semi}. Consequently, we employ a graph U-Net structure \cite{gao2019graph} which is illustrated in Figure~\ref{fig:graphUnet}. This module has an encoder-decoder structure \cite{ronneberger2015u} with a number of skip connections that concatenate the encoders and decoder features, along with a number of GCN layers (depicted in Figure ~\ref{fig:graphUnet} as {\em Graph conv} layers). 

For a GCN layer, we define a graph $\textit{G} = ( {N}, {A} )$, where ${N}$ is the number of nodes and ${A} \in \mathbb{R}^{{N} \times {N}}$ as the adjacency matrix. The values of the adjacency matrix are defined based on the relationship between nodes; if two nodes are connected, the value is equal to 1, otherwise equal to 0. In HPE, one often applies the kinematic structure of the hand skeleton as the adjacency matrix \cite{chen2021mvhm,Ge_2019_CVPR,cai2019exploiting}. In this paper, however, we learn the adjacency matrix to allow for more advanced connections to be dynamically discovered automatically.
This approach was proposed by \cite{doosti2020hope}, and our results (presented in the next section) demonstrate that this approach in fact boosts performance when compared to random as well as pre-defined adjacency matrices.


The output of the layer with 
$\textit{F}$ input features
and trainable weight matrix ${W} \in \mathbb{R}^{{N} \times {L}}$, where ${L}$ is the output feature size, is computed according to \begin{equation}
\textit{Y} = \sigma(\bar{{A}}{X}{W}),
\end{equation}
where ${X} \in \mathbb{R}^{{N} \times {F}}$ is the GCN layer input and $\bar{\textit{A}}$ is the normalized adjacency matrix of the graph \cite{kipf2016semi}. $\bar{\textit{A}}$ is measured as
\begin{equation}
\bar{{A}} = {D}^{\frac{-1}{2}}\hat{{A}}{D}^{\frac{-1}{2}},
\end{equation}
where 
\begin{equation}
\hat{{A}} = {A} + {I},
\end{equation}
${D}$ is the diagonal node degree matrix, and ${I}$ is the identity matrix. Accordingly, the graph U-Net module ${G}_{v}^ {t}$ transforms each 2D coordinate ${w}_{v}^ {t}$ to the 3D camera coordinate ${p}_{v}^ {t}$ at time $t$ and view $v$ such that
\begin{equation}
{p}_{v}^ {t} = {G}_{v}^ {t}({w}_{v}^ {t}),
{p}\in  \mathbb{R}^{21 \times 3},
{w}\in  \mathbb{R}^{21 \times 2} .
\end{equation}

\subsection{Training and Implementation Details} \label{sec:Implementation}

We employ a multi-stage training strategy for MuViHandNet. First (Stage 1)
we aim to train the pipeline irrespective of the temporal and sequential learner components. To this end, we temporarily replace the LSTM networks with a fully connected layer and train the entire pipeline, essentially re-training the image encoder (which is ResNet-10 pre-trained by ImageNet \cite{russakovsky2015imagenet}) and training the graph U-Net. Next (Stage 2), we replace the temporary FC layer with the original LSTMs and retrain the entire network while the image encoder is kept frozen, in essence training the temporal and sequential learners and re-training the $T \times V$ instances of the graph U-Net. 

\begin{table}[!t]

\caption{Training hyper-parameters used in Stages 1 and 2.}
\setlength
\tabcolsep{4pt}
\centering
\begin{tabular}
    {l| l| l| l}
    \hline
    \textbf{Sub-Net} & \textbf{Parameter} & \textbf{Stage 1}  & \textbf{Stage 2}\\
    \hline
    \hline
  Encoder &Architecture &ResNet-10  &ResNet-10\\
    &Pretrained &ImageNet  &Stage 1\\
     &\# of Inputs &1 &15 \\
  &Embedding Layer & Avg. Pooling  &Avg. Pooling \\
  & Feature Size &512&512 \\
    \hline
    Multi-view & \# of Inputs &- &3\\
    LSTM            & \# of hidden layers &- &2\\
                  & Hidden Size  &- &128\\
    \hline
    Temporal  & \# of Inputs &-  &5\\
     LSTM            & \# of hidden layers  &-  &2\\
                  & Hidden Size  &- &128\\
                              \hline
    FC + ReLU   &Dimensionality &256 &-\\ 
\hline
    FC + ReLU   &Dimensionality &128 &128\\
    \hline
    FC + ReLU &Dimensionality &$21\times2$ &$21\times2$\\
    \hline
        Drop out &Size &-&0.25\\
    \hline
    Graph U-Net& \# of modules &1&15\\
                &Output size &$21\times3$  &$21\times3$\\
    \hline
    Full Network &Batch Size  &64 &8\\
    &Loss Function &$0.01L_{2D} + L_{3D}$ &$L_{3D}$\\
    &Optimizer&Adam &Adam\\
    &Learning rate &0.001 &0.006\\
    &Weight Decay &0.1&0.07\\
    &Step Decay &100 &100\\
    &\# of Epochs &500  &400\\ 
\hline
    \end{tabular}
    \label{tab:model_details}
\end{table}

In stage 1, we train the image encoder and the graph U-Net for 500 epochs with Adam optimizer. An initial learning rate of 0.001 is used and multiplied by 0.1 every 100 epochs. The utilized loss function is
\begin{equation}
     L = \alpha L_{2D} + L_{3D}.
\end{equation}
Here, the $L_{2D}$ is calculated according to 
\begin{equation}
    L_{2D} =||\hat{w} - w||_2,
\end{equation}
where $\hat{w}$ and $w$ are the predicted and ground truth 2D coordinates respectively. Also, the $L_{3D}$ is measured according to 
\begin{equation}\label{eq:3d_loss}
    L_{3D} =||\hat{p} - p||_2,
\end{equation}
where $\hat{p}$ and $p$ are the predicted and ground truth 3D coordinates, respectively. 

In stage 2, when the temporal and angular learners are added to the pipeline, training is performed for 400 epochs. Here, the 
loss function is 
\begin{equation}
    L = \frac{1}{{V\times T}}\sum_{t}\sum_{v}||\hat{p}_{v}^ {t} - p_{v}^ {t} ||_2, 
\end{equation}
where $\hat{p}_{v}^ {t}$ and $p_{v}^ {t}$ are respectively the predicted and ground truth 3D coordinates at time $t$ and view $v$. Table \ref{tab:model_details} summarizes the hyper parameters used for two stages of training. In our implementation the video length $T$ is equal to 5 and the number of the views $V$ is equal to 3. Our implementation has been done in PyTorch, using an Nvidia GeForce GTX 2070 Ti GPU.

\begin{table*}[!t]
\caption{Performance comparison between MuViHandNet and prior works with cross-subject and cross-activity testing protocols. 
}
    \centering\begin{tabular}
    {|c| l | c |l |c |c|c|c|}
    \hline
    Test &Method & Encoder &Loss Function & Root Pos. &$\downarrow$ Avg. EPE & $\downarrow$ Avg. median EPE & $\uparrow$AUC\\
    \hline
    \hline
    {\multirow{11}{*}{\rotatebox[origin=c]{90}{cross-subject}}}
     &Boukhayma et al. \cite{boukhayma20193d} &ResNet-50 &$L_{mask}, L_{2D},L_{3D} ,  L_{\theta }$ &Yes&135.899 &138.276 &0.012\\
    &Boukhayma et al. \cite{boukhayma20193d} &ResNet-50 &$L_{3D}$  &Yes & 48.840  &40.837  &0.280\\
     &Hasson et al. \cite{hasson2019learning} &ResNet-18 & $L_{3D},L_{\beta}$ &Yes &62.674  &65.524 &0.18\\
     &Hasson et al. \cite{hasson2019learning} &ResNet-18 & $L_{3D}$ &Yes&28.915  &24.764  &0.574\\
     &Doosti et al. \cite{doosti2020hope}  &ResNet-10 & $L_{init2D}, L_{2D}, L_{3D}$&No& 18.895  &16.635  & 0.634\\
    
    &MuViHandNet(GRU$_v$,GRU$_t$) & ResNet-10 & $L_{2D},  L_{3D}$&No &13.450  &11.646 & 0.739\\
    &MuViHandNet(LSTM$_v$,GRU$_t$) & ResNet-10 & $L_{2D},  L_{3D}$&No &10.493 &8.715 &0.798\\
    &MuViHandNet(LSTM$_t$,GRU$_v$) & ResNet-10 & $L_{2D},  L_{3D}$&No &10.092 &{8.380} &{0.807}\\
     &MuViHandNet(AutoEnc) & ResNet-10 &  $L_{2D},  L_{3D}$&No &27.461   & 24.828  &  0.480\\
     &MuViHandNet(GCN) & ResNet-10 & $L_{2D},  L_{3D}$&No &13.175  &11.263 &0.745
\\
    & \textbf{MuViHandNet (proposed)} & ResNet-10 & $L_{2D},  L_{3D}$&No &\textbf{8.881} &\textbf{7.351} &\textbf{0.831}\\

    \hline
    {\multirow{9}{*}{\rotatebox[origin=c]{90}{cross-activity}}}
    &Boukhayma t al.\cite{boukhayma20193d} &ResNet-50 &$L_{3D}$  &Yes &42.799 &40.238 &0.287\\
    &Hasson et al. \cite{hasson2019learning} &ResNet-18 & $L_{3D}$ &Yes&66.851 & 68.457   & 0.152\\
    &Doosti et al. \cite{doosti2020hope}  &ResNet-10 & $L_{init2D}, L_{2D}, L_{3D}$&No& 46.745 &45.086  &  0.217 \\
   
    &MuViHandNet(GRU$_v$,GRU$_t$) & ResNet-10 & $L_{2D},  L_{3D}$&No &22.065  & 20.726 & 0.575\\
    &MuViHandNet(LSTM$_v$,GRU$_t$) & ResNet-10 & $L_{2D},  L_{3D}$&No & 21.266 &19.186 & 0.589\\
    &MuViHandNet(LSTM$_t$,GRU$_v$) & ResNet-10 & $L_{2D},  L_{3D}$&No &23.222& 21.241& 0.553\\
     &MuViHandNet(AutoEnc) & ResNet-10 & $L_{2D},  L_{3D}$&No &31.212  &30.812&0.423\\
    &MuViHandNet(GCN) & ResNet-10 & $L_{2D},  L_{3D}$&No &29.506  & 27.957 & 0.446 \\
    &\textbf{MuViHandNet (proposed)} & ResNet-10 & $L_{2D},  L_{3D}$&No &\textbf{20.375} &\textbf{17.819} &\textbf{0.608}\\
    \hline
    \end{tabular}
    \label{tab:otherMethod}
\end{table*}

\begin{figure}[!t]
\centerline{\includegraphics[width=0.4\textwidth]{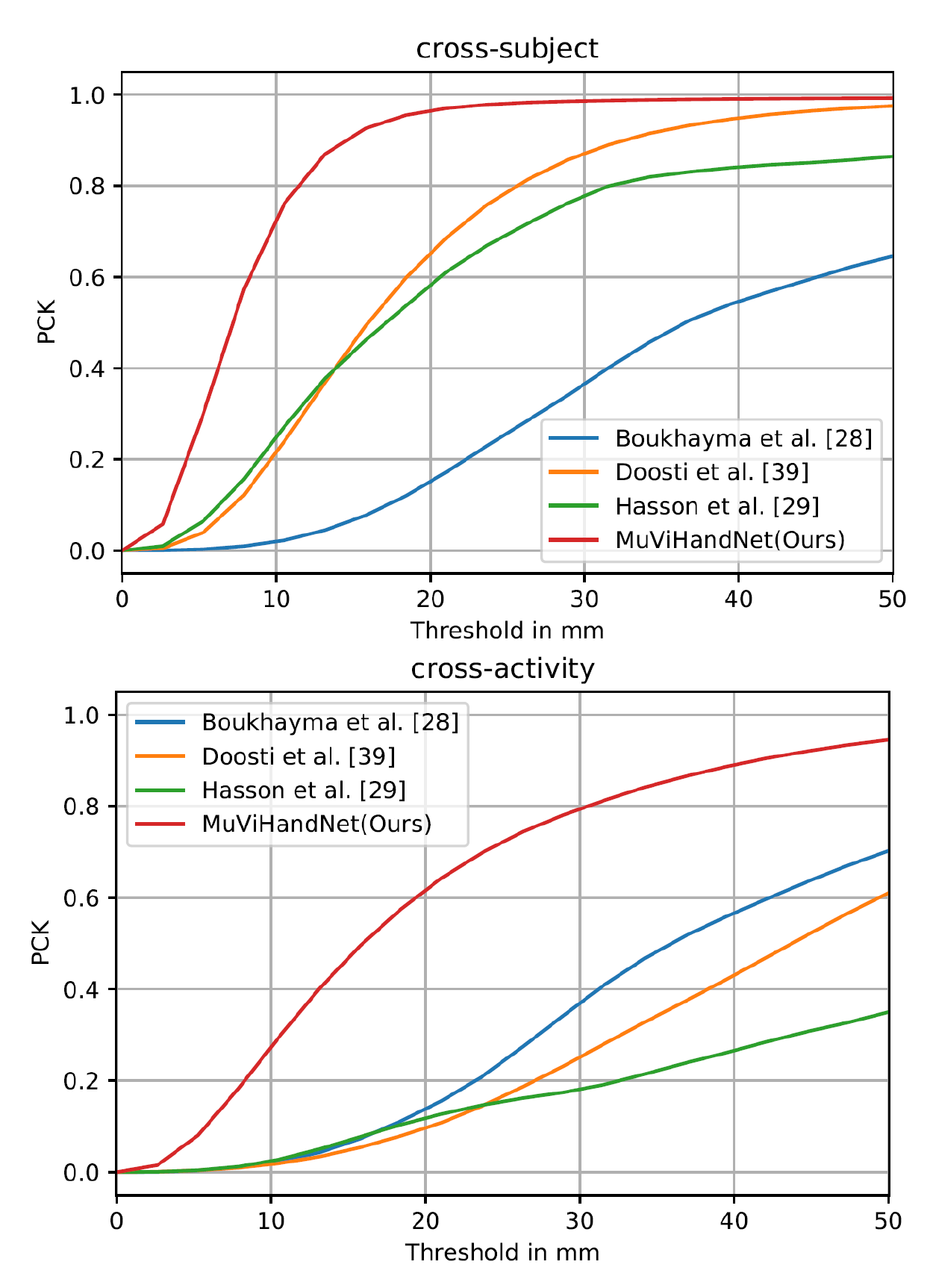}}
\caption {Comparison of PCK curves for our method in comparison with state-of-the-art solutions \cite{hasson2019learning,boukhayma20193d,doosti2020hope}.
Two different testing protocols have been used: cross-subject (top) and cross-activity (bottom).}
\label{fig:SOT_curve}
\end{figure}

\section{Experiments and Results}

In this section, we describe our experiments and report on the results. We also report on the outcome of ablation experiments and investigate the effects of various components of our network on the overall HPE performance.

\subsection{Test Protocol and Evaluation}\label{sec:Test Protocol}
To rigorously evaluate the result of our proposed method, two evaluation protocols have been tested on the MuViHand dataset, \textit{i)} \textit{cross-subject}, in which seven subjects are used for training the network (subjects $\{= 3, \cdots, 9\}$) and three other subjects (subjects $ = \{1, 2, 10\}$) with different variations of skin tones, appearance, and gender are set aside for testing; \textit{ii)} \textit{cross-activity} in which two random activities (activities $= \{8, 19\}$) are taken for testing and 17 activities (activities $= \{1,\cdots, 7, 9, \cdots, 18\}$) are used in the training phase. In this research, similar to previous studies \cite{zimmermann2017learning,hasson2019learning}, we focus on estimating the hand pose from one hand only (right hand) from the cropped hand images. Nonetheless, our dataset allows for future work to focus on the other hand or HPE from non-cropped fully body images. Accordingly, we utilize the images captured from Cameras 7 to 9 in the MuViHand dataset, in which the right hand images are at the center of the frames and they have been resized to $224\times 224$ pixels to fit the input size of the image encoder. 


Similar to \cite{zimmermann2017learning} for evaluating our method, three metrics are used. These include (\textit{i}) the percentage of correct key points (PCK) with a threshold between 0-50 mm; (\textit{ii}) the area under the curve (AUC) of the PCK; (\textit{iii}) and the mean and median endpoint error (EPE).

\begin{figure*}[!t]
\centerline{\includegraphics[width=0.95\textwidth]{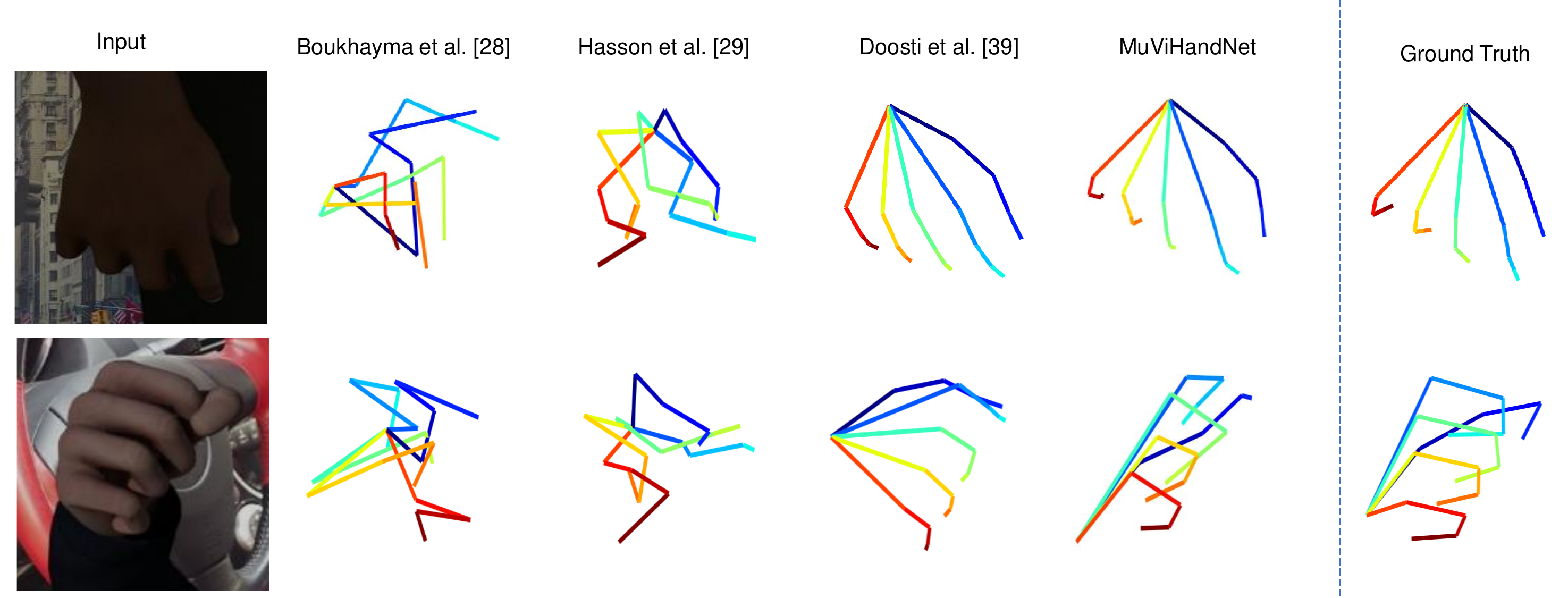}}
\caption {Samples of estimated hand poses using our method in comparison with other works \cite{hasson2019learning,boukhayma20193d,doosti2020hope} are presented.}
\label{fig:sample3}
\end{figure*}

\subsection{Benchmarking Methods}\label{sec:Benchmarking}
We compare our results against three state-of-the-art 3D HPE methods \cite{hasson2019learning,boukhayma20193d,doosti2020hope}. These works have been selected as benchmarks for the following reasons: (1) they obtained very strong results for 3D HPE; (2) their implementations are publicly available, which is essential given that the benchmarks need to be re-trained on our newly proposed dataset; (3) similar to our proposed method, they do not use any additional modalities such as depth towards 3D HPE. Nevertheless, because these methods have been originally optimized for other datasets and not MuViHand, we tune the parameters of the three state-of-the-art methods to obtain the best possible performance to allow for a fair comparison.

In addition to state-of-the-art benchmarks, we create and evaluate several variations of the proposed MuViHandNet. First, we substitute the LSTM temporal and angular learners with Gated recurrent unit (GRU) networks. We refer to this variation of the model as MuViHandNet(GRU$_v$,GRU$_t$). Next, we explore using a combination of LSTM and GRU for temporal and angular learners, and vice versa. These are referred to a MuViHandNet(LSTM$_v$,GRU$_t$) and  MuViHandNet(GRU$_v$,LSTM$_t$). We then create another benchmark variant by swapping the GCN layers of the graph U-Net structure with fully connected layers, essentially creating a stacked autoencoder. We refer to this variant as MuViHandNet(AutoEnc). Lastly, we modify the graph U-Net to no longer have a U-Net architecture by using a three GCN layers instead. This variant is referred to as MuViHandNet(GCN). To differentiate the original model as proposed in Section \ref{sec:Method}, we use the term MuViHandNet(proposed) in the tables in Section \ref{sec:Method}.

\begin{table}[!t]
\caption{Impact of window size which is used in the LSTM$_t$ of the MuViHandNet on the EPE. }
\setlength
\tabcolsep{4pt}
    \centering\begin{tabular}{|l|*{5}{c}|}\hline
\backslashbox{Test}{Win. Size}
&\makebox{3}&\makebox{5}&\makebox{7}
&\makebox{9}&\makebox{11}\\\hline\hline
cross-subject & 11.197 &\textbf{8.881} & 13.432 &12.149  &14.710 \\\hline
cross-activity & 21.605 &\textbf{20.375} &24.621 &22.991  & 27.290\\\hline
\end{tabular}
  \label{tab:window_size}
\end{table}

\begin{table}[!t]
\caption{The impact of different types of adjacency matrices in our graph U-Net.}
\setlength
\tabcolsep{3pt}
    \centering\begin{tabular}{|l|*{5}{c}|}\hline
\backslashbox{Test}{Adj. Mat.}
&\makebox{Rand 1}&\makebox{Rand 2}&\makebox{Rand 3}
&\makebox{Hand Skel.}&\makebox{Learned}\\\hline\hline
cross-subject & 20.951 &15.136 & 43.225& 16.951 &\textbf{8.881}  \\
\hline
cross-activity & 25.829 & 22.826 & 23.565 &21.354 &\textbf{20.375} \\
\hline
\end{tabular}
  \label{tab:adjacency}
\end{table}

\begin{table*}[!t]
\centering
\caption{Ablation studies on different components of MuViHand.  A breakdown of results based on joints and fingers is also provided. }
    
    \setlength
    \tabcolsep{3pt}
    
    \begin{tabular}
    {| c | c | c c |c c| c c c c c |c c c c c |}
    \hline
     Test &Method & LSTM$_t$  &LSTM$_v$  & $\downarrow$ EPE & $\uparrow$ AUC &$\downarrow$ Wrist &$\downarrow$ MCP &$\downarrow$ PIP &$\downarrow$ DIP  &$\downarrow$TIP &$\downarrow$Thumb &$\downarrow$Index &$\downarrow$Middle &$\downarrow$Ring &$\downarrow$Pinkie \\
  
    \hline
     \hline
      {\multirow{4}{*}{\rotatebox[origin=c]{90}{cr.-sub.}}}

     &Baseline 1 & \xmark &\cmark &10.034  &0.808 &43.316 &8.389 &8.413 &9.644 &12.929&10.829 &10.6986 &9.252 &8.530 &9.910 \\
     &Baseline 2&\cmark& \xmark  &11.823 &0.766 &53.096 &9.329 &9.622 &11.617 &15.924 &13.274 &12.226 &10.968 &9.966 &11.682\\
     &Baseline 3 & \xmark & \xmark  &14.529  &0.723 &61.555 &9.852 &11.8118 &15.030 &19.928 &15.388 &14.728 &13.281 &12.809 &14.572\\     
     &Full Model & \cmark & \cmark &\textbf{8.881} &\textbf{0.831} &\textbf{36.882} &\textbf{7.277} &\textbf{7.7330} &\textbf{8.583} &\textbf{11.086}&\textbf{9.220} &\textbf{8.8473} &\textbf{7.773} &\textbf{7.820} &\textbf{9.687}  \\
     \hline
      {\multirow{4}{*}{\rotatebox[origin=c]{90}{cr.-act.}}}
    
     &Baseline 1 & \xmark & \cmark & 21.463 &0.592 &107.105 &14.327 &15.300 &20.837 &33.241  &26.776 &21.185 &17.660 &18.451 &20.559\\
     &Baseline 2 &\cmark & \xmark&23.631&0.557 &120.143 & 16.129 &19.197 & 27.238 &40.598  &26.776 &21.185 &17.660 &18.451 &20.559\\
     &Baseline 3 & \xmark & \xmark  &44.021 &0.246 &177.371 &29.700 &39.055 &50.392 &63.873 &44.342 &48.227 &44.054 &43.4031 &48.749 \\
     &Full Model  & \cmark & \cmark &\textbf{20.375}  &\textbf{0.608} &\textbf{100.462} &\textbf{13.605} &\textbf{14.586} &\textbf{20.188} &\textbf{31.6684}  &\textbf{25.115} &\textbf{21.135} &\textbf{16.636} &\textbf{17.332} &\textbf{19.640}\\
     \hline
    \end{tabular}
    \label{tab:result_abiliation}
\end{table*}

\begin{figure}[!t]
\centerline{\includegraphics[width=0.5\textwidth]{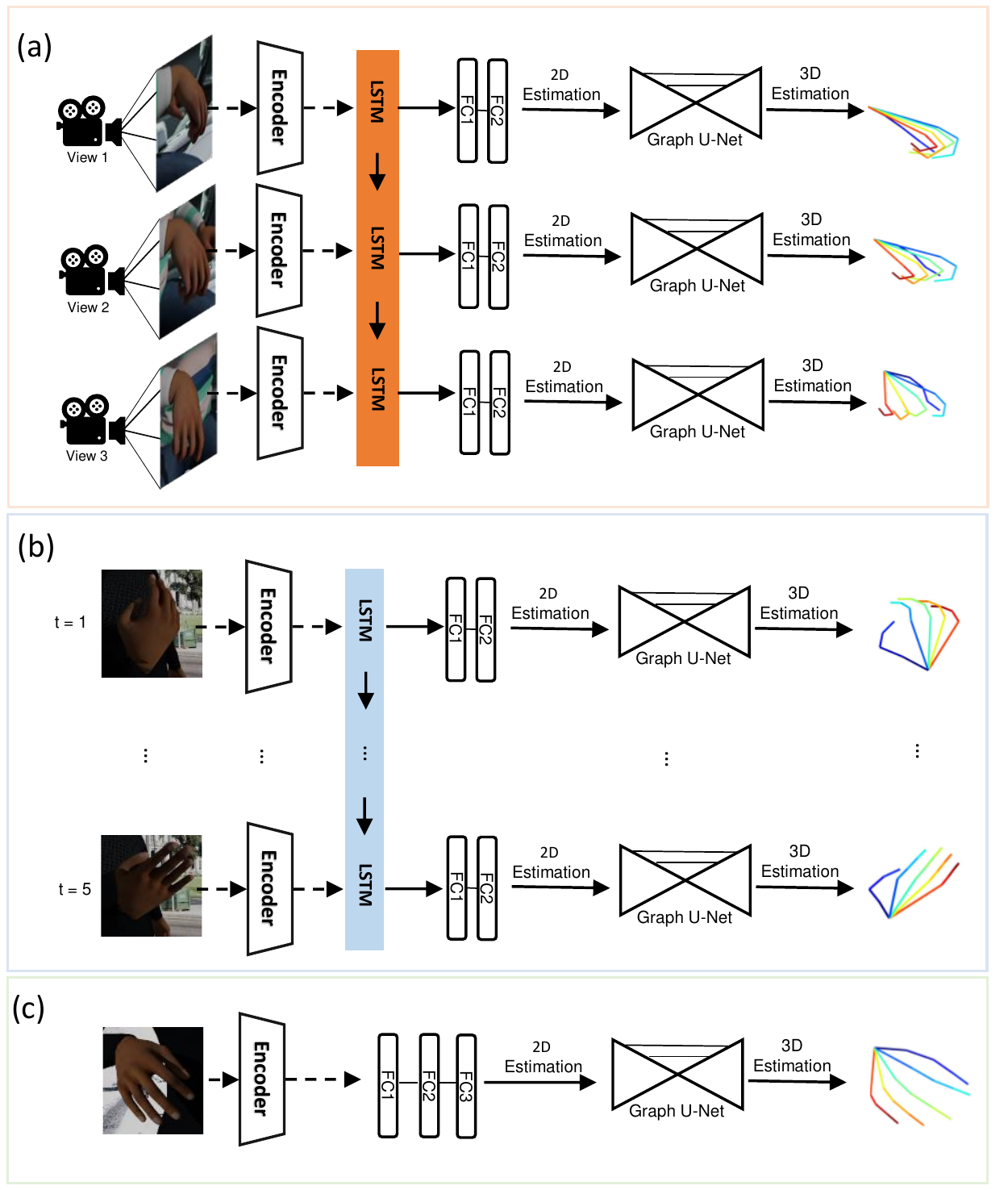}}
\caption{The network architectures for ablated variations of MuViHandNet: (a) Baseline 1, (b) Baseline 2, and (c) Baseline 3.}
\label{fig:LSTM_mv model}
\end{figure}

\begin{figure}[!t]
\centerline{\includegraphics[width=0.4\textwidth]{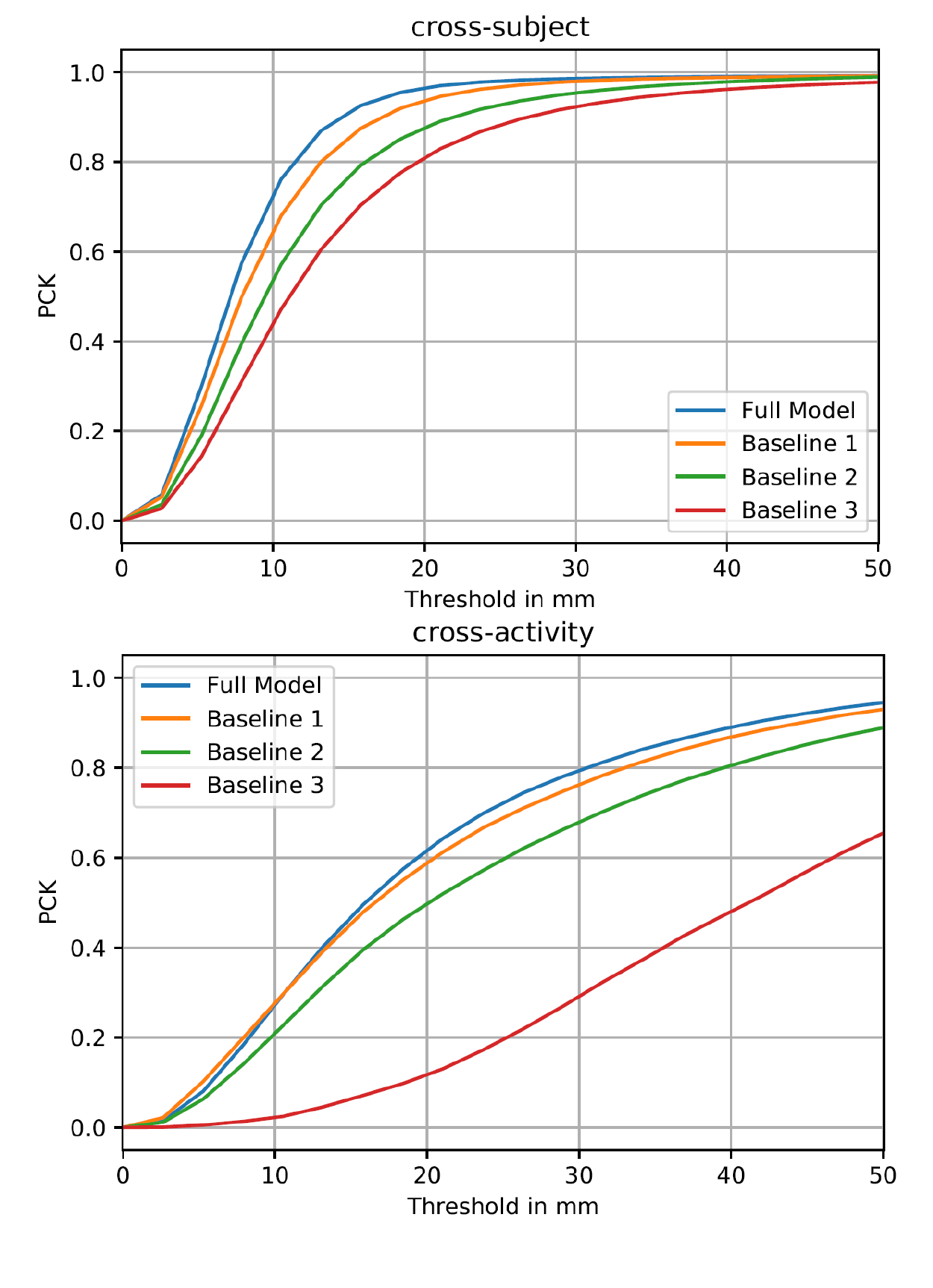}}
\caption {Comparison of PCK curves for 3 ablated variations of MuViHandNet along with the full model. cross-subject (top) and cross-activity (bottom).}
\label{fig:CURVEs}
\end{figure}

\subsection{Performance and Discussion}

The performance of MuViHandNet in comparison with the state-of-the-art methods \cite{hasson2019learning,boukhayma20193d,doosti2020hope}, along with the variations discussed in Section \ref{sec:Benchmarking} for the two test protocols described in Section \ref{sec:Test Protocol}, are presented in Table \ref{tab:otherMethod}.  MuViHandNet outperforms the state-of-the-art methods (with different loss functions) in both testing protocols by a considerable margin. It is also evident that the proposed method including the graph U-Net and LSTM learners outperforms the benchmarking variants, which we described in Section \ref{sec:Benchmarking}. When comparing the performance of MuViHandNet in the cross-subject scheme with cross-activity, the cross-activity protocol is far more challenging. This is in line with prior works that have shown that HPE methods often fail on the unseen poses \cite{armagan2020measuring}. Moreover, most of the prior works \cite{hasson2019learning,boukhayma20193d} require the root pose as an input to these models, while \cite{doosti2020hope}, along with our method, operate without such input. 

Figure \ref{fig:SOT_curve} presents the PCK curves for various thresholds (0-50) for the state-of-the-art methods for the two test protocols,
as well as the proposed MuViHandNet. To subjectively evaluate the performance of our method, we highlight two challenging images along with the detected poses by our method along with the state-of-the-art benchmarks\cite{hasson2019learning,boukhayma20193d,doosti2020hope}. Note that existing methods often perform poorly when dealing with such challenging scenarios where lighting conditions are relatively poor or the pose contains hidden hand parts and fingers. This points to (a) the effectiveness of our proposed method, and (b) the challenging nature and thus contribution of our proposed dataset.

To evaluate the impact of the selected temporal window size (number of frames), we perform an experiment that involved changing this parameter (3, 5, 7, 9, 11) and the number of cells in the temporal learner of MuViHandNet. The results of this experiment are presented in Table~\ref{tab:window_size}, where we observe that a window size of 5 yields the best results.

As discussed earlier in Section~\ref{sec:graohUnet} one of the advantages of our work is the integration of an adjacency matrix for the graph U-Net, which can be learned through the network as opposed to the common approach of pre-defining this matrix. To evaluate the impact of this approach, we compare the use of random as well as predefined adjacency matrices with our learned method. 
The predefined baseline adjacency matrix is defined based on the skeletal architecture of the hand, similar to \cite{doosti2020hope}. The results are presented in Table~\ref{tab:adjacency}, where our strategy shows the best performance.


\begin{figure*}[!t]
\centerline{\includegraphics[width=0.95\textwidth]{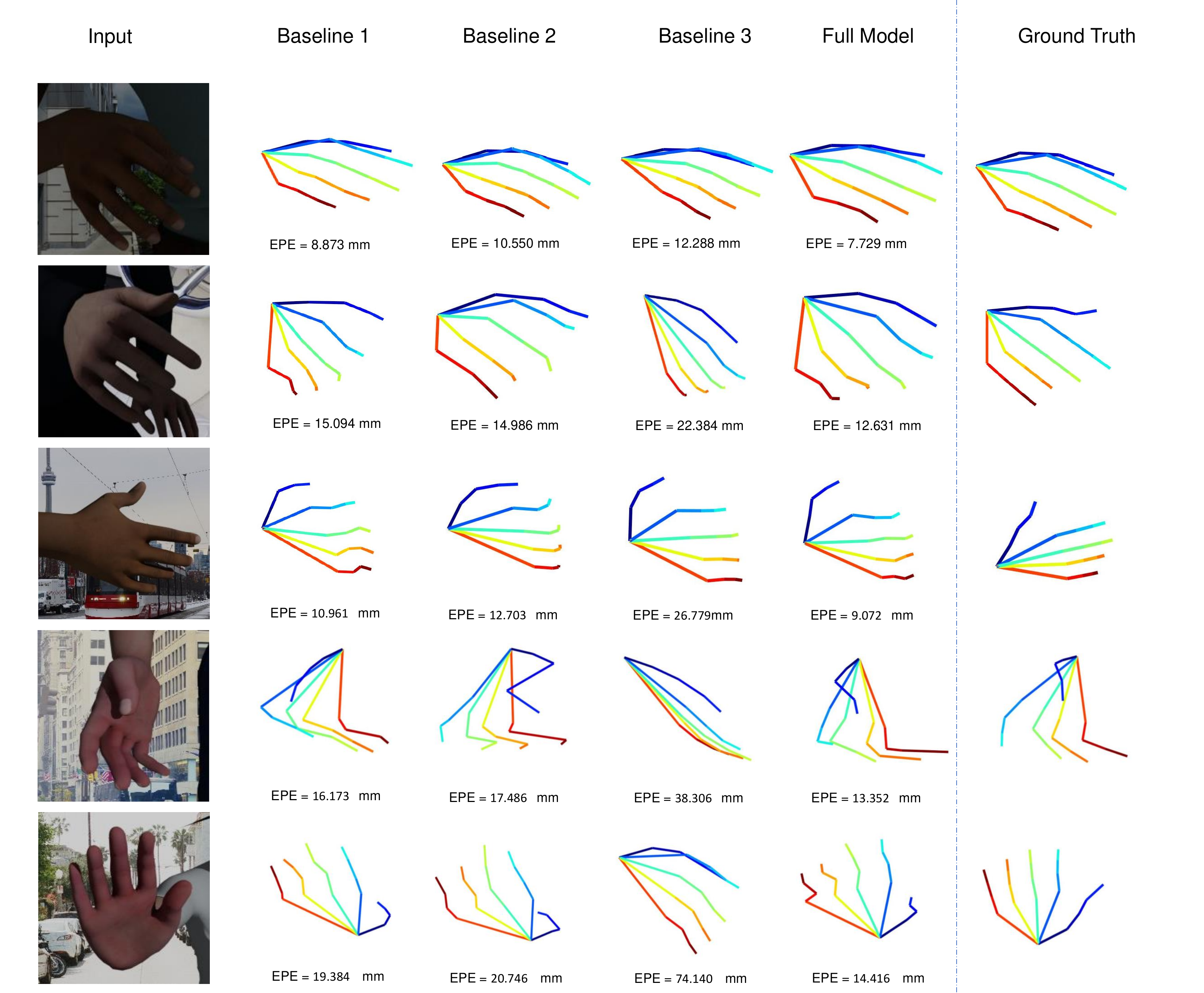}}
\caption {Sample visualizations of estimated hand poses using MuViHandNet and its ablated variations under challenging environmental conditions. In particular, row 1 deals with poor illumination, row 2 deals with a challenging background (same color tone as the hand), and rows 3 through 5 deal with different viewing angles. with no self-occlusion.}
\label{fig:sample1}
\end{figure*}

\begin{figure*}[!t]
\centerline{\includegraphics[width=0.95\textwidth]{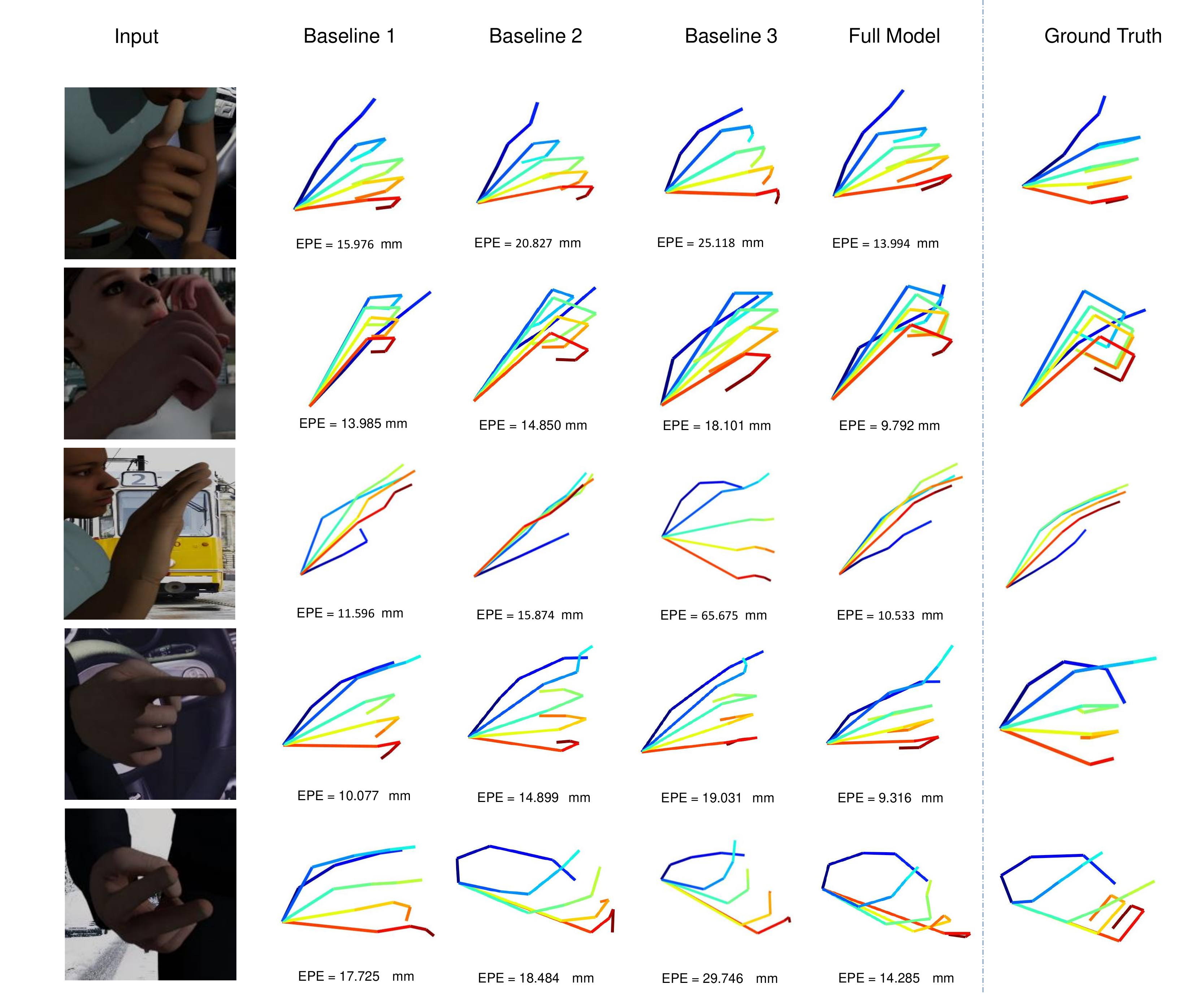}}
\caption {Sample visualizations of estimated hand poses with self-occlusion using MuViHandNet and its ablated variations.}
\label{fig:sample2}
\end{figure*}

\begin{figure}[!t]
\centerline{\includegraphics[width=0.4\textwidth]{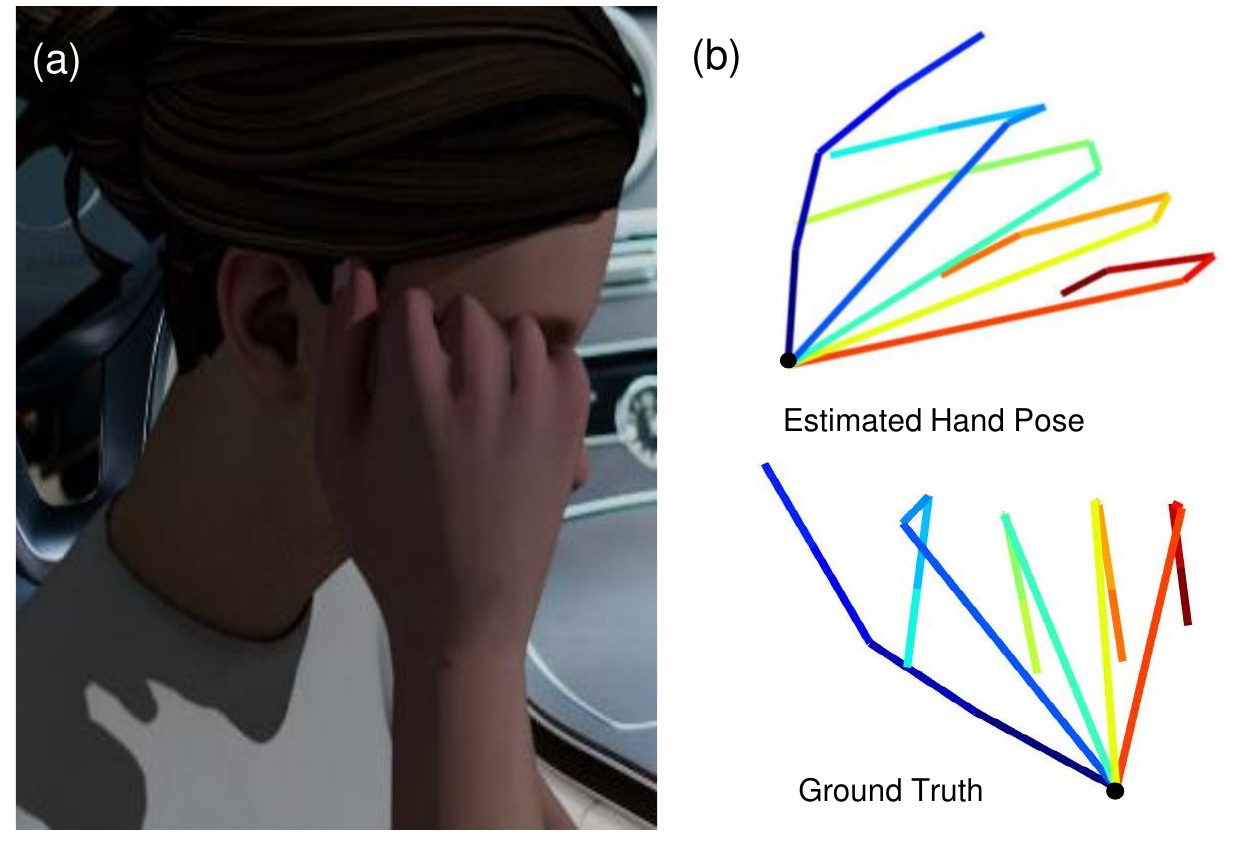}}
\caption{ (a) A sample challenging frame from the dataset where a difficult background (skin) and self-occlusions are seen; (b) The hand pose estimated by MuViHandNet along with the ground truth are presented where MuViHandNet performs poorly. }
\label{fig:fail}
\end{figure}

\subsection{Ablation Study}
To explore the impact of each component of MuViHandNet on our results, we create three ablated variations by systematically removing the main components of the model as follows. First, we remove the temporal learner module, LSTM$_t$. The ablated model is depicted in Figure \ref{fig:LSTM_mv model}(a). In this model, the encoder accepts frames from three different views and generates a 512 dimensional image embedding ${F_v}, v = 1,2,3$ for each view. The $[F_1, F_2,F_3] \in \mathbb{R}^{3 \times 512}$ is then fed to the angular learner LSTM$_v$ to generate a feature vector $Y_v, v = 1,2,3$, where each ${Y_v}$ is fed to two FC layers for generating the 2D coordinates followed by the graph U-Net to transform the 2D joint locations to 3D camera coordinates for each hand frame. Next, we remove the angular learner LSTM$_v$ from the structure of MuViHandNet, as shown in Figure \ref{fig:LSTM_mv model}(b). This method takes five consecutive frames and, similar to the previous ablation experiment, a 512 dimensional image embedding $F_t, t  = 1,2, \cdots ,5$ is generated for each frame to then feed the rest of the network. Lastly, both sequential learners LSTM$_t$ and LSTM$_v$ are removed, meaning that only a single hand frame is passed through the encoder followed by 3 FC layers and a graph U-Net module to estimate the 3D hand pose, as shown in Figure \ref{fig:LSTM_mv model}(c). 

The results of this experiment are presented in Table \ref{tab:result_abiliation} for both testing protocols (cross-subject and cross-activity).  The EPE is broken out across each joint and finger. These results demonstrate the advantages of each of the two angular and temporal learners in our pipeline. Here, when the angular learner is removed, the EPE increases by around 1.3 mm and 1.2 mm for cross-subject and cross-activity schemes, respectively. Next, when the angular learner is removed, the EPE increases by approximately 3 mm and 3.5 mm in the cross-subject and cross-activity schemes.  This illustrates that the impact of learning angular information is higher than that of temporal information for 3D HPE. Finally, when both are removed, our method suffers from an increase in EPE of around 6 mm and 24 mm for the two evaluation schemes, respectively. This also demonstrates the added value of having both temporal and angular information in our dataset.

By breaking down the EPE across each joint, we remark that removal of both temporal and angular learners significantly increases the EPE for the wrist, pointing that the inclusion of the sequential learners in our model considerably impacts the ability to locate the position of the hand. Finally, when we break down the ablation experiments for different fingers, we observe that the removal of both sequential learners impacts the pinkie finger more negatively than the others, which could be due to the higher likelihood for this finger to be obstructed by other fingers. The table presents the AUC, which also shows similar trends for all of the above-mentioned experiments. 

Lastly, Figure \ref{fig:CURVEs} shows the PCK curves for our ablation study, where consistent behaviours for different threshold values are observed. 
Further, we visualize the estimated hand poses for several images with no self-occlusion using MuViHandNet and its ablated variations in Figure \ref{fig:sample1}. It is observed that MuViHandNet performs better under poor illumination (row 1), challenging background with the same color tone as the hand (row 2), and different viewing angles (rows 3 through 5). Additionally, we present the performance of our method on samples with severe self-occlusions in Figure \ref{fig:sample2}. Here, we observe that in these difficult scenarios, the performance drops considerably for the three baseline models in comparison to MuViHandNet. This is because our proposed pipeline can effectively learn information from additional views (angular leaner) or frames (temporal learner) to obtain a better sense of the occluded joints.





\subsection{Limitations and Future Work}

The MuViHand dataset includes full-body images of subjects, and thus often depicts the hands in  front of other body parts, which generally contain the same skin tone. 
As expectd, such scenarios posed challenges for MuViHandNet (see Figure \ref{fig:fail}) and while our model still performed better than other existing methods, the results could be further improved. Adding further modalities such as depth can allow for the model to focus on the hand in the foreground to overcome such issues, which we can explore in future work.

Moreover, our dataset was developed by synthetic images. For future work, to allow for more in-the-wild applicability, the use of generative adversarial networks might be explored to add realism to the synthetic images with the help of a real hand images. Lastly, weakly supervised techniques could be used to combine our dataset (which contains accurate ground-truths) with real-world datasets (which often do not contain accurate ground-truths) for HPE applications.

\section{Conclusion}

In this research we proposed a novel multi-view video-based hand pose datasets consisting of synthetic images and ground-truth 2D/3D pose values. Our dataset, MuViHand includes more than 402,000 images in 4,560 videos, and has been captured from six different views with 12 cameras in two concentric circles (one fixed and the to track the hand). The dataset, which we make public, is the first to include synthetic videos in a multi-view setting and provides a rich resource for performing 3D HPE in challenging scenarios. Next, we proposed a new model, MuViHandNet for detection HPE on our dataset. The model consists of different components for encoding images, learning temporal and angular relationships, and estimating the 3D poses. We performed rigorous experiments to evaluate the performance of our model, including comparisons with other methods and a number of ablated baselines. Our experiments demonstrated the effectiveness of our method as well as the challenging nature of our developed dataset.



\bibliographystyle{unsrt}  
\bibliography{IEEE}

\ifCLASSOPTIONcaptionsoff
  \newpage
\fi

\end{document}